\documentclass[lettersize,journal]{IEEEtran}
\usepackage[labelformat=simple, labelfont={small,bf}]{subcaption}
\usepackage{amsmath,amsfonts}
\usepackage{algorithmic}
\usepackage{algorithm}
\usepackage{array}
\usepackage{textcomp}
\usepackage{stfloats}
\usepackage{url}
\usepackage{verbatim}
\usepackage{graphicx}
\usepackage{cite}
\usepackage{hyperref}
\usepackage[labelformat=simple, labelfont={small,bf}]{subcaption}
\usepackage{todonotes}
\usepackage{booktabs}
\RequirePackage{xstring}
\RequirePackage{xparse}
\RequirePackage[]{acro}

\makeatletter
\@ifpackagelater{acro}{2019/02/17}{
    \@ifpackagelater{acro}{2020/04/29}{
        \NewDocumentCommand\acrodef{mO{#1}mG{}}{\DeclareAcronym{#1}{short={#2}, long={#3}, #4}}
    }{
        \NewDocumentCommand\acrodef{mO{#1}mG{}}{\DeclareAcronym{#1}{short={#2}, long={#3}, foreign-plural={}, #4}}
    }
}{
    \NewDocumentCommand\acrodef{mO{#1}mG{}}{\DeclareAcronym{#1}{short={#2}, long={#3}, #4}}
}
\makeatother
\makeatletter
\@ifpackagelater{acro}{2020/04/29}{
    \NewAcroTemplate{inparen}{%
        \acroiffirstTF{%
            \acrowrite{long}%
            \acspace%
            \scalebox{0.5}[1]{(}%
            \acroifT{foreign}{\acrowrite{foreign}, }%
            \acrowrite{short}%
            \acroifT{alt}{ \acrotranslate{or} \acrowrite{alt}}%
            \acrogroupcite
            \scalebox{0.5}[1]{)}%
        }{%
            \acrowrite{short}%
        }%
    }
    \NewAcroTemplate{paren}{%
        \acroiffirstTF{%
            \acrowrite{long}%
            ,%
            \acspace%
            \acroifT{foreign}{\acrowrite{foreign}, }%
            \acrowrite{short}%
            \acroifT{alt}{ \acrotranslate{or} \acrowrite{alt}}%
            \acrogroupcite
        }{%
            \acrowrite{short}%
        }%
    }
}{
    \DeclareAcroFirstStyle{inparen}{inline}
      {
        brackets = true,
        brackets-type = {%
          \scalebox{0.5}[1]{(}%
        }{%
          \scalebox{0.5}[1]{)}%
        }
      }
}
\makeatother

\acrodef{ADAS}{Advanced Driver Assistance Systems}
\acrodef{AV}{Autonomous Vehicle}
\acrodef{BDI}{Belief-Desire-Intention}
\acrodef{CAM}{Cooperative Awareness Message}
\acrodef{C-ITS}{Cooperative Intelligent Transport Systems}
\acrodef{CINEA}{European Climate, Infrastructure and Environment Executive Agency}
\acrodef{CPM}{Collective Perception Message}
\acrodef{CPU}{Central Processing Unit}
\acrodef{DENM}{Decentralized Environmental Notification Message}
\acrodef{DoF}{Degree of Freedom}
\acrodef{ETSI}{European Telecommunications Standards Institute}
\acrodef{FSM}{Finite State Machine}
\acrodef{GPS}{Global Positioning System}
\acrodef{HRI}{Human-Robot Interaction}
\acrodef{IMU}{Inertial Measurement Unit}
\acrodef{IPQ}{Igroup Presence Questionnaire}
\acrodef{ITQ}{Immersive Tendencies Questionnaire}
\acrodef{IVIM}{In-Vehicle Information Message}
\acrodef{KIT}{Karlsruhe Institute of Technology}
\acrodef{LCD}{Liquid Crystal Display}
\acrodef{LTE-V2X}{Long Term Evolution Vehicle-to-Everything}
\acrodef{MAPEM}{MAP Extended Message}
\acrodef{MDP}{Markov Decision Process}
\acrodef{NARS}{Negative Attitudes Towards Robots Scale}
\acrodef{NR-V2X}{New Radio Vehicle-to-Everything}
\acrodef{OBU}{On-Board Unit}
\acrodef{PoC}{Proof of Concept}
\acrodef{RAS}{Robotic Awareness Service}
\acrodef{RMSSD}{Root Mean Square of Successive Differences}
\acrodef{RSU}{Roadside Unit}
\acrodef{SARAI}{Socially Assistive Robotics with Artificial Intelligence}
\acrodef{SDNN}{Standard Deviation of NN intervals}
\acrodef{SPATEM}{Signal Phase and Timing Extended Message}
\acrodef{SREM}{Signal Request Extended Message}
\acrodef{SSEM}{Signal Status Extended Message}
\acrodef{SSQ}{Simulator Sickness Questionnaire}
\acrodef{UMLSL}{Urban Multi-lane Spatial Logic}
\acrodef{V2X}{Vehicle-to-Everything}
\acrodef{VR}{Virtual Reality}
\acrodef{VRU}{Vulnerable Road User}
\acrodef{ZoD}{Zone of Danger}

\begin{document}

\title{ROBOPOL: Social Robotics Meets Vehicular Communications for Cooperative Automated Driving}

\author{John Pravin Arockiasamy, Andy Comeca, Victoria Yang, Manuel Bied, Maximilian Schrapel, Alexey Rolich, \\ Barbara Bruno, Maike Schwammberger, Dieter Fiems and Alexey Vinel
\thanks{J.~P.~Arockiasamy, A.~Comeca, V.~Yang, M.~Bied, M.~Schrapel, B.~Bruno, M.~Schwammberger and A.~Vinel are with Karlsruhe Institute of Technology (KIT), Germany; A.~Rolich is with the University of Roma "Sapienza", Italy; D.~Fiems is with Ghent Univerity, Belgium.}
\thanks{A.Vinel is a corresponding author, email:  \href{mailto:alexey.vinel@kit.edu}{alexey.vinel@kit.edu}.}
\thanks{This work has been funded as part of the KIT Future Fields Stage 2 project "V2X4Robot", the Excellence Strategy of the German Federal and State Governments, and the Helmholtz Program “Engineering Digital Futures”. This work is also part of the "CulturalRoad" project, funded by the European Union under grant agreement No. 101147397. Views and opinions expressed are however those of the authors only and do not necessarily reflect those of the European Union or the European Climate, Infrastructure and Environment Executive Agency (CINEA). Neither the European Union nor the granting authority can be held responsible for them.
}
\thanks{Manuscript revised April 02, 2026.}}



\maketitle

\begin{abstract}
On the way toward full autonomy, sharing roads between automated and autonomous vehicles in so-called mixed traffic is unavoidable. Moreover, even if all vehicles on the road were autonomous, pedestrians would still cross streets. We propose social robots as moderators between autonomous vehicles and vulnerable road users. This paper presents a first proof-of-concept integration of a social robot advising pedestrians in crossing scenarios involving a cooperative automated vehicle. We also discuss key enablers required for designing “robot policeman” in a generic use case of cooperative intersection management. Our work provides a vision of the role of social robotics in future Cooperative Intelligent Transport Systems.
\end{abstract}



\section{Introduction}
\IEEEPARstart{O}n the way towards a ubiquitous proliferation of \acp{AV}, intermediate automation adoption phases are practically unavoidable. While \acp{AV} fully take over driving responsibilities, \emph{automated} vehicles can assist human drivers through \ac{ADAS}. Furthermore, there is gradual progress in \ac{V2X} communications and respective \emph{cooperation} abilities of the vehicles. Since both these processes will not happen homogeneously for all vehicles at once, different vehicles with different levels of autonomy and cooperation will share the space on our roads~\cite{Richard2023} in a \emph{mixed traffic}. Classifications reflecting such technological deployment phases consider both SAE-defined automation levels (Level 0 to Level 5) and \ac{V2X} connectivity phases (Day 1, Day 2, Day 3+). The highest level of autonomy envisions vehicles operating independently, while the lower levels require varying degrees of human intervention. At lower levels, human drivers primarily operate vehicles, progressing to \acp{AV} driving themselves with occasional human input. \ac{V2X} phases follow a similar progression: Day 1 focuses on basic status updates performed by the vehicles, Day 2 enables sensor data sharing, and Day 3+ incorporates cooperative maneuvering capabilities enabled by exchange and adjustment of their trajectories to prioritize efficiency and safety~\cite{Ott2022}.


Even if over time the automation and connectivity of vehicles increases, this does not necessarily apply to \acp{VRU}. 
As long as traffic will also include human participants, especially \acp{VRU}, such as pedestrians and cyclists, a true global increase in safety and efficiency is only possible if humans are allowed to partake in and benefit from the information sharing. In other words, the information generated and communicated by vehicular systems needs to be available to humans, in human-understandable ways~\cite{bied2024autonomous}.

Current solutions for traffic regulation are either envisioned for human or automated traffic actors, but rarely both: \acp{AV} can only communicate with other \acp{AV}, while human policemen can effectively communicate with humans, but not with \acp{AV}. 
We argue that traffic regulation, especially considering the foreseen expected spread of \acp{AV}, is a task that can benefit from the introduction of technologies enabling a \emph{seamless communication between autonomous and human traffic actors}. 
\begin{figure}[!t]
\centering
\includegraphics[width=.7\linewidth]{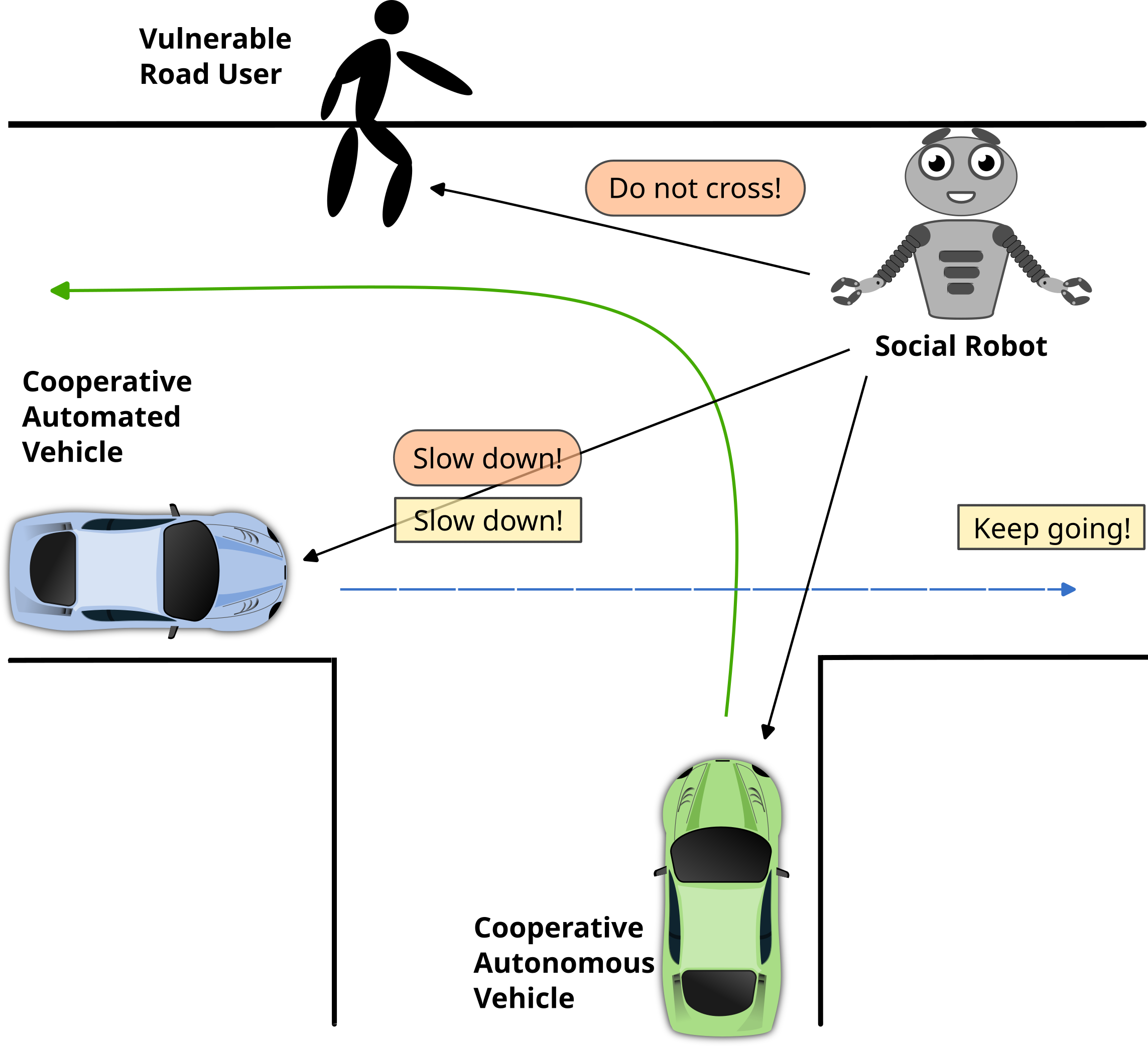}
\caption{Cooperative social robot communicating with both \acp{VRU} and \acp{AV} to coordinate the traffic using natural (orange boxes with round edges) as well as digital communication (yellow boxes with sharp edges).}
\label{V2X4Robot}
\vspace{-0.5cm}
\end{figure}
In this paper, we address this challenge by investigating the feasibility of \emph{social robots as traffic mediators} in mixed-traffic scenarios~\cite{MuC}. We postulate that the introduction of social robots on the roads can integrate human and/or non-\ac{V2X}-enabled actors in cooperative intelligent transportation systems (C-ITS). Such an approach allows relaxing an impractical requirement that all human participants, including \ac{VRU} such as pedestrians and cyclists, of all ages, are \ac{V2X}-capable all the time\cite{Oscar2022}. The implication that every human traffic actor (1) owns a \ac{V2X}-capable device, (2) knows how to use it and (3) always has it on and with them is practically impossible to guarantee in practice. Certainly, the assumption can be partly addressed via a collective perception approach~\cite{Delooz2022}, where the information about \acp{VRU} locally sensed by cooperative vehicles (or the infrastructure) is made available to technical actors using \ac{V2X}. Nevertheless, \acp{VRU} tracking via collective perception and even their full \ac{V2X}-instrumentation primarily enable a mono-directional information flow, and e.g. limit safety-related solutions to those that exclusively rely on actions taken by the technical actors. 
Smartphone vibration or alarming sounds can be easily ignored or misinterpreted and, as they are not currently associated with traffic management, would require learning and adaptation by each and every human traffic participant~\cite{Andre2024}.

This is where we foresee the potential for social robots, that can concurrently communicate both with \ac{V2X}-enabled vehicles and with humans, even conveying traffic-related information and instructions in ways that are familiar to humans, e.g., mimicking the gestures used by human policemen (therefore, we refer to our solution as \emph{ROBot POLiceman} or \emph{ROBOPOL}), or the light patterns of traffic lights.
The presence of such a communication mediator is even more important in cases of disruptions or breakdowns of the standard traffic regulation infrastructure, where the robot could be deployed to support the work of human policemen (also ensuring that they can devote most of their time and attention to those tasks that require their specific training in de-escalating dangerous situations). 


The remainder of this manuscript is organized as follows. Section~II provides an overview of related work. Section~III motivates the use case, while Section~IV presents the corresponding proof-of-concept (PoC). The main technical enablers of ROBOPOL in the broader context of cooperative intersection management are discussed in Section~V. Finally, Section~VI concludes the paper.

\section{Related Work}


The idea of using a robot in traffic has recently appeared in the literature. For example, \cite{lee2015study} conducts a survey on user preferences of a robot that measures the speed of vehicles and sends warnings to road workers compared to a lane changing board and a speeding sign board, finding that the robot was favored for safety and convenience, but lower on economic viability. In \cite{mirnigThreeStrategiesAutonomous2017}, the idea of a robot as a familiar proxy for communication between AV and pedestrians is briefly introduced, but neither elaborated nor tested. However, this idea is tested in virtual reality by \cite{hollanderInvestigatingInfluenceExternal2019} where a gesturing robotic driver is displayed on the windshield. The virtual robotic driver is communicating to pedestrians to compensate for the absence of the driver in a (simulated) autonomous vehicle. In \cite{ghaffarControllingTrafficHumanoid}, another traffic robot that uses gestures to control traffic is presented, but again only tested in simulation. Moreover, an experimental setup for a user study aiming to evaluate the appearance of the used robot is presented. However, since the quantitative study was only conducted with two participants, it is not clear how to interpret the results.   
Similarly, \cite{najjar2022towards} presents a robot model for a traffic signal management system in which the robot is placed at the intersection of a road network in a web-based traffic simulator. The focus here is on vehicles, and \acp{VRU} are not considered. 
Some of the works that include an implementation of real robots exclusively focus on the perception of and/or the communication with some traffic actors.
In \cite{kumaranCrossNotToCrossRobotic}, a robot is placed on the dashboard of a vehicle that appears to be autonomous to communicate to pedestrians if it's safe to cross in front of the vehicle or not. The system is however only tested with a stationary vehicle and the robot is designed to exclusively convey information to nearby pedestrians when wirelessly triggered by a human operator. Similarly, \cite{gong2017real} presents a humanoid robot that can be remotely controlled and uses gestures to communicate with road users. In this case, the robot does not use its own perception and does not interact fully autonomously, but relies on the human operator by using human-in-the-loop control \cite{daniHumanLoopRobotControl2020}.  

As the whole research field of Human-Robot Interaction (HRI) \cite{bartneck2024human} reveals, robot characteristics such as its embodiment play a crucial role in the naturalness and effectiveness of human-robot communication \cite{roeslerMetaanalysisEffectivenessAnthropomorphism2021}. For example, \cite{chen2020traffic} proposes a guide-dog robot to assist vision-impaired individuals in determining when to cross the street, which again exclusively focuses on the user-robot communication and relies on the robot's own visual capabilities to extract relevant information about the environment. 

In \cite{FloresComeca2025SocialRF} and \cite{Comeca2025RobotsFS}, a robot traffic management system is implemented as non-\ac{V2X} PoC solutions, focusing on the use of sensors onboard for vehicle and pedestrian detection and information processing. However, hazard detection relies completely on onboard sensor perception, which is limited by the field of view, not taking into account vehicles coming from non-visible areas for the robot. 

To the best of our knowledge, the only project that integrates \ac{V2X} capability with a traffic robot is the \emph{IPA2X project}\footnote{https://ipa2x.eu/}. Complementary to this, in \cite{Chauhan2026b}, a mobile Smart Pole Interaction Unit (SPIU) has been investigated as a pedestrian-side mediator that provides explicit “WALK/STOP” cues and coordinates them with a vehicle-mounted interface. Here, a \ac{V2X}-enabled robot is used to help children, the elderly, and people with disabilities to safely cross the street. In these works, a real robot has been deployed that relies on \ac{V2X} communication. However, the interaction with the \acp{VRU} only includes simple visual and acoustic signals and does not include more complex human-robot interaction. In such settings, a similar crossing-guard function could in principle be implemented without \ac{V2X}, for example by relying solely on on-board perception and explicit stop/go signalling as in \cite{crossbot}. However, purely line-of-sight sensing remains susceptible to occlusions and blind spots at crossings, which collective perception via \ac{V2X} aims to mitigate~\cite{occlusionproblem}. 

In summary, none of the existing works identifies the key enablers required to successfully deploy a \ac{V2X}-enabled robot capable of orchestrating mixed traffic, nor do they implement all of these enablers in a PoC demonstrator. Our work aims to fill this gap and opens new opportunities for future research on a wide range of ROBOPOL use cases and implementations.

\section{First steps}


\subsection{Choice of a use-case}
Traffic environments are becoming increasingly complex due to the coexistence of connected vehicles, automated systems, cyclists, and pedestrians. While existing infrastructure such as pedestrian crossings, road side units and traffic lights has proven effective in conventional settings, it faces limitations in mixed traffic scenarios involving automated and connected vehicles. One important question to answer is how pedestrians can proportionally benefit from technological advances in vehicular technology. This is due to two main challenges in recent developments. First, pedestrians remain largely excluded from \ac{V2X} communication, resulting in a lack of transparency regarding vehicle intent and system state \cite{morales2025, bied2024autonomous}. This can lead to uncertainty, reduced trust, and potentially unsafe interactions, especially in unregulated or weakly regulated crossing situations. Second, while \acp{AV} promise various benefits as comfort, safety and efficiency \cite{litman2023autonomous,winkleSafetyBenefitsAutomated2016}, they lack the possibility for the pedestrian to communicate with the human driver via social cues. A majority of pedestrians seem to communicate their intention to cross via body posture and gaze~\cite{rasouliAgreeingCrossHow2017, suchaPedestriandriverCommunicationDecision2017} with the goal to evoke compliance with the driver and search for acknowledgment that they were seen~\cite{rothenbucherGhostDriverField2016}. Therefore, the lack of a human driver deprives pedestrians of an important communication strategy with human drivers \cite{bied2024autonomous}.

To address these challenges, namely giving pedestrians the possibility to benefit from \ac{V2X} technology, and the reduction of social interaction in mixed traffic, we introduce a social robot equipped with \ac{V2X} capability as an intermediary that facilitates communication between connected vehicles and pedestrians. The robot leverages \ac{V2X} communication to receive real-time information from nearby vehicles and translates this information into human-understandable signals. Compared to traditional infrastructure such as traffic lights and road side units, a robotic system offers several advantages: it is mobile, can position itself dynamically where needed (including temporary or degraded situations), and can actively perceive and interact with nearby \acp{VRU}. This enables interaction in close proximity to the \acp{VRU} which is likely to increase its persuasiveness~\cite{liBenefitBeingPhysically2015a}.  Furthermore, social robots allow for richer and understandable communication modalities including the possibility of multi-modal communication~\cite{bairy2026}. This communication can benefit from the human tendencies to attribute social characteristics such as intelligence or likeability to robotic systems \cite{Bartneck2009}. While it remains an open research question how these attributes can be effectively leveraged in traffic contexts, they provide a promising foundation for designing more expressive and adaptive interaction strategies than static signaling systems \cite{bied2024autonomous}. Solutions like human traffic officers equipped with \ac{V2X} might prove efficient in some context, however they do not fully solve these challenges. While human traffic officers are inherently better suited for social interaction than social robots, they are arguably not as well suited to process large amounts of \ac{V2X} data as robots. Furthermore, using a social robot instead of traffic officers reduces the risk for humans. Simpler solutions like a static road side unit equipped with an external HMI could be possible for some context.  However, they lack the mobility and the capability to adapt their positioning w.r.t. pedestrians.    

In this PoC, we investigate a \emph{pedestrian crossing assistance} without traffic lights, where pedestrians must independently assess safe crossing opportunities. The PoC is performed in a pedestrian-priority area where pedestrians legally have the right of way. While the robot’s behavior to physically blocking or instructing pedestrians to stop, does not fully align with the expected traffic dynamics in such zones, this discrepancy is negligible in the context of this PoC. In practice, the pedestrians usually wait for approaching vehicles to pass before they cross. Furthermore, this location was chosen for safety and legal reasons, since the car speed is reasonably low and the area is managed by KIT. A connected e-bike is used as an approaching vehicle, broadcasting its status via \ac{V2X} messages; it was selected for pragmatic reasons, including cost and safety considerations. The robot is positioned near the roadside, where it detects pedestrians and their intent to cross while simultaneously processing incoming \ac{V2X} data. By combining these inputs, the robot estimates the current traffic situation and determines whether it is safe to cross. If a crossing is deemed unsafe, the robot signals the pedestrian to wait; once conditions are safe, it indicates that crossing is permitted. Through this behavior, the robot acts as a traffic mediator, demonstrating how robotic systems can enhance transparency and support decision-making for pedestrians in connected traffic environments.

\subsection{Pre-study in a virtual reality}

Virtual Reality (VR) enables immersive experiences to evaluate future robot use cases in traffic scenarios without compromising safety~\cite{argota2024virtual}. Although VR has limitations in terms of realism and field of view, it remains a powerful tool to evaluate safety-critical scenarios with humans in the loop without any harm. We use this technology as an early stage design loop to iteratively refine the social robot's communication cues and appearance for pedestrian interactions in pedestrian crossing scenarios before field deployment. To minimize development effort and enhance realism, we present 360-degree video recordings from a pedestrian point of view  through a VR headset. The videos capture a diverse set of traffic scenarios across multiple environments, with varying traffic density and road user constellations. Additional recorded road crossings without the social robot serve as a baseline reference to assess how the robot's presence and its cues influence pedestrian behavior across the different environments. 

To quantify how the robot’s presence and communication cues affect pedestrians beyond qualitative feedback, we capture subjective, behavioral, and physiological measures during the VR exposure. Perceived safety is a commonly used outcome in VR traffic research to capture subjective comfort and predictability in relation to vehicle motion and scene structure~\cite{argota2024virtual, lee2021perceived,walker2019feeling}. We therefore collect a scene-wise perceived safety rating after each clip on a continuous 0-100 scale. As the head pose and turning behavior near the curb provides informative cues about pedestrian uncertainty and crossing intention~\cite{rasouli2017they}, we log head orientation streams at 30~Hz from the headset and derive indicators of visual scanning and attention allocation, such as forward oriented gaze share, yaw rotation intensity, and micro and macro turns. We also record heart rate at 130~Hz and its variability using a Polar H10 chest strap and compute standard time-domain indices such as RMSSD, SDNN, as well as stress-related indices including the Baevsky stress index~\cite{shaffer2017overview}. Validated questionnaires complement our study design to control for VR-related confounds and individual differences, including SSQ~\cite{kennedy1993simulator} and IPQ~\cite{schubert2001experience} for motion sickness and presence in virtual realities, ITQ and ATI~\cite{franke2019personal} for immersive tendency and technology affinity, as well as Godspeed~\cite{Bartneck2009} and NARS~\cite{nomura2006experimental} for robot perception and attitudes.

To complement the controlled measurements, we gathered early qualitative feedback during an exploratory public demonstration for a scientific Human-Computer Interaction audience at a conference~\cite{MuC}. The demonstration combined the physical robot interaction with immersive 360 degree VR videos shown on external screens and through a Meta Quest 3 headset. Around 50 attendees visited the demo, and about 20 experienced the VR scenario. Although the demonstration was not a formally controlled study, the feedback was largely positive. Participants particularly commented on robot height, facial expressiveness, and the importance of aligning gesture timing with traffic flow and pedestrian actions. They also suggested additional settings for future exploration, including school crossings, parking lots, and pedestrian-tram interactions in public space. No participant reported motion sickness. Overall, the demonstration provided valuable exploratory input for refining the robot design and extending the concept to further traffic scenarios.

\subsection{General Finite State Machine}
\label{sec:fsm_logic}
\begin{figure}[!htb]
\centering
\includegraphics[width=.95\linewidth]{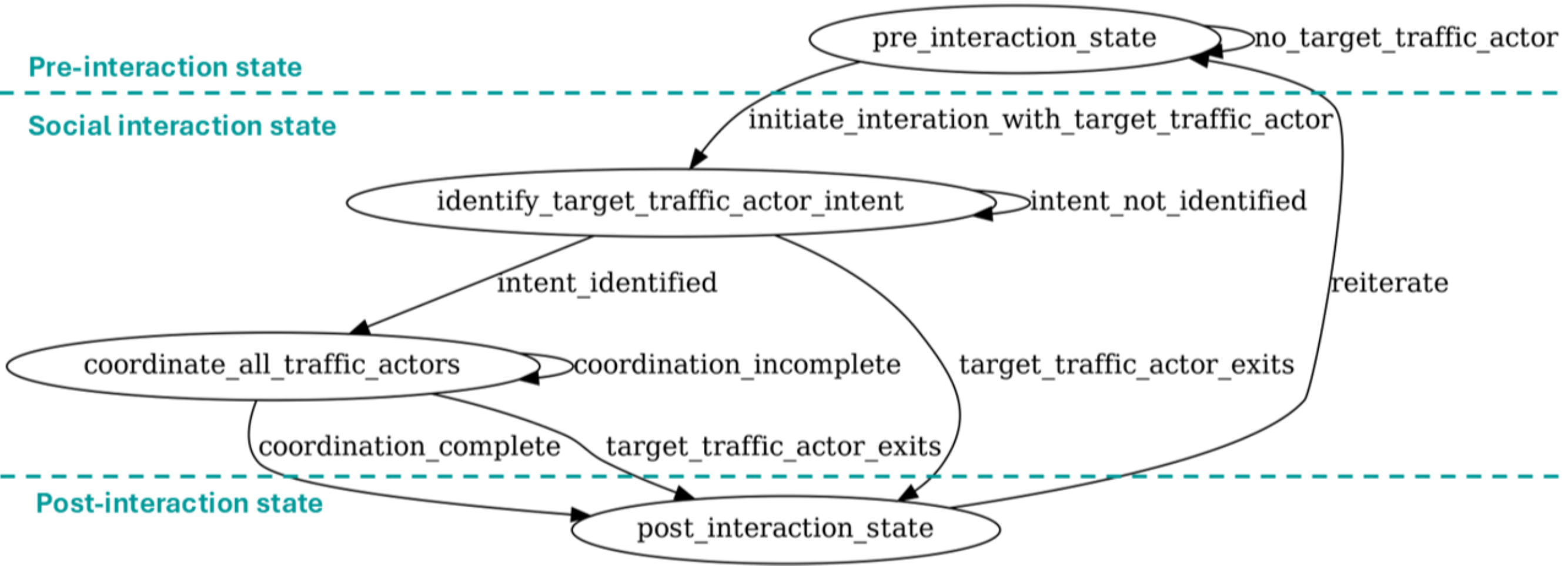}
\caption{Generalized \ac{FSM} that can be applied to different use cases and be centric to any traffic actor with \ac{V2X} capability.}
\label{fig:state_machine_general}
\vspace{-0.5cm}
\end{figure}

As a result of our initial pre-study, we created a generalized \ac{FSM} formalizing ROBOPOL operation (Fig. \ref{fig:state_machine_general}) that can be applied to any use case and be centric to any \ac{V2X}-enabled traffic actor. The states of the \ac{FSM} were modeled in terms of different stages of an interaction (i.e., involving communication and coordination) between all traffic actors to ultimately achieve the goal of the target traffic actor. There are three main parts in our \ac{FSM}, 1) pre-interaction, 2) social interaction, and 3) post-interaction. The \ac{FSM} is organized in terms of stages of a social interaction, because the focus is on a social human-robot interaction that leverages information from \ac{V2X} communication to benefit \acp{VRU} in the form in introducing AV technology and mixed traffic regulation and mediation. The pre-interaction state entails the identification of the target traffic actor, done by the advanced perception. The social interaction state entails the intent identification of target traffic actor, coordination of all traffic actors to achieve an ideal environment for the target traffic actor’s intent, and the execution of the planned. This is done by \ac{V2X}, social HRI, and formal specification of driving maneuvers, respectively. The post-interaction state entails the system reiterating and learning from the outcomes of the interaction. In this state parameters may be updated such that the next iteration can perform better during the pre-interaction and social interaction states. For the general state machine, the term ``traffic actors" is used to imply that the general state machine can be used for any traffic entity (for a vehicle, bike, pedestrian, driver, etc). The ``target traffic actor" is the entity that the system is focused on helping, e.g., in the crossing use case, the target traffic actor is the pedestrian, but in other cases this can be a vehicle, or anything else.  

\section{Proof-of-concept}
\subsection{Our equipment}
We selected PAL Robotics-ARI\footnote{https://pal-robotics.com/robot/ari/}, a humanoid robot standing at 165 cm tall, designed to facilitate natural human-robot interactions, as the ROBOPOL platform. With its advanced suite of sensors and actuators, ARI is well-suited to perceive and respond to its environment in real time. This makes it a compelling choice for managing traffic scenarios, helping pedestrians cross streets, and interacting with autonomous vehicles. 

ARI’s sensory capabilities encompasses one RGB-D cameras positioned in the front head, which is used to track pedestrians in a 10m range, that allow the robot to anticipate the pedestrian presence according to our scenario (see Fig. \ref{fig:poc_top_view}). ARI is also equipped with a two-degree-of-freedom (DoF) head for movement, that increase the field of view of the robot and makes it more interactive. Additionally, it uses a differential mobile base with two DoFs for navigation purposes in case the robot needs to move to the road, and two robotic arms, each with five DoFs, to allow the execution of a wide range of gestures to convey intuitive signals and foster engaging interactions with pedestrians and drivers. 

\begin{figure}[t]
\centering
\includegraphics[width=.8\linewidth]{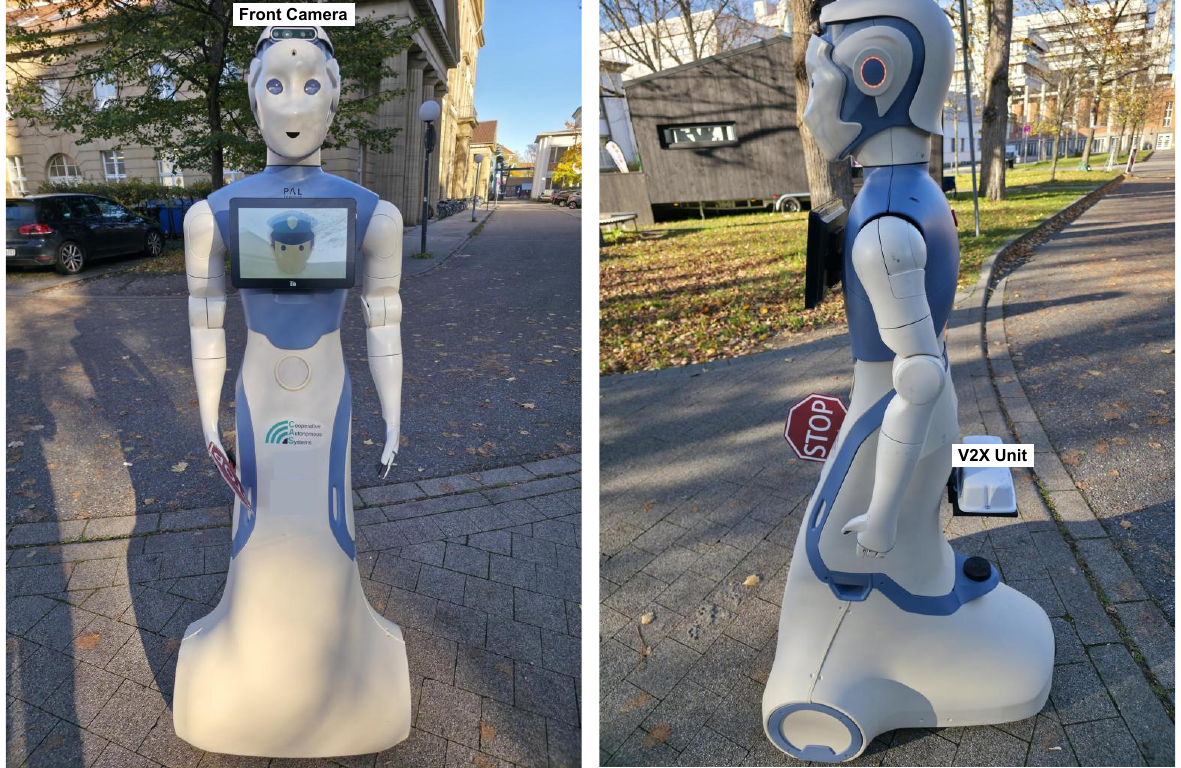}
\caption{Front and side views of the ARI, equipped with a rear-mounted \ac{V2X} unit, and a front-facing RGB-D camera.}
\label{fig:ari_hardware}
\vspace{-0.3cm}
\end{figure}

For \ac{V2X} communication, we integrated a \ac{V2X} On-Board Unit (OBU) (UCU 5.0) from Herman\footnote{https://www.herman.cz/en/} onto ARI as illustrated in Fig. \ref{fig:ari_hardware}. This OBU complies with \ac{ETSI} ITS G5 standards and facilitates the transmission and reception of \ac{CAM}, \ac{DENM}, IVIM, MAPEM, SPATEM, SREM, and SSEM messages. It includes GPS and IMU sensors with a 400-meter range in urban areas, allowing ARI to exchange critical traffic information with \ac{V2X}-enabled vehicles. 

\begin{figure}[t]
\centering
\includegraphics[width=.7\linewidth]{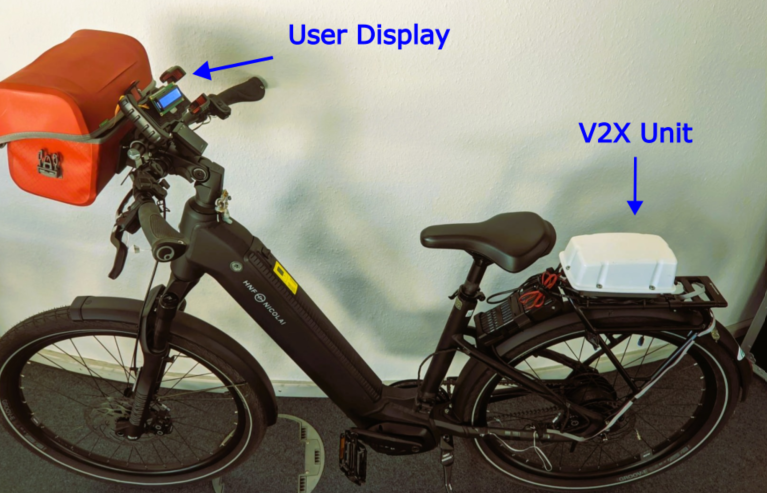}
\caption{The HNF Nicolai UD4 All-Terrain e-bike equipped with a \ac{V2X} communication unit and a display interface for potential hazard warnings}
\label{fig:bike_hardware}
\vspace{-0.5cm}
\end{figure}

To simulate vehicle interaction, we used an HNF Nicolai UD4 All-Terrain e-bike equipped with the same \ac{V2X} OBU, as depicted in Fig.
\ref{fig:bike_hardware}. Additionally, the e-bike is equipped with a 16×2 character LCD display connected to a Raspberry Pi that is used to present any warning messages received by the OBUs. This setup ensures safe testing while demonstrating ARI's \ac{V2X} communication capabilities. However, any vehicle equipped with \ac{V2X} OBU is suitable for this PoC.

\subsection{System architecture}
\begin{figure}[!htb]
\centering
\includegraphics[width=.8\linewidth]{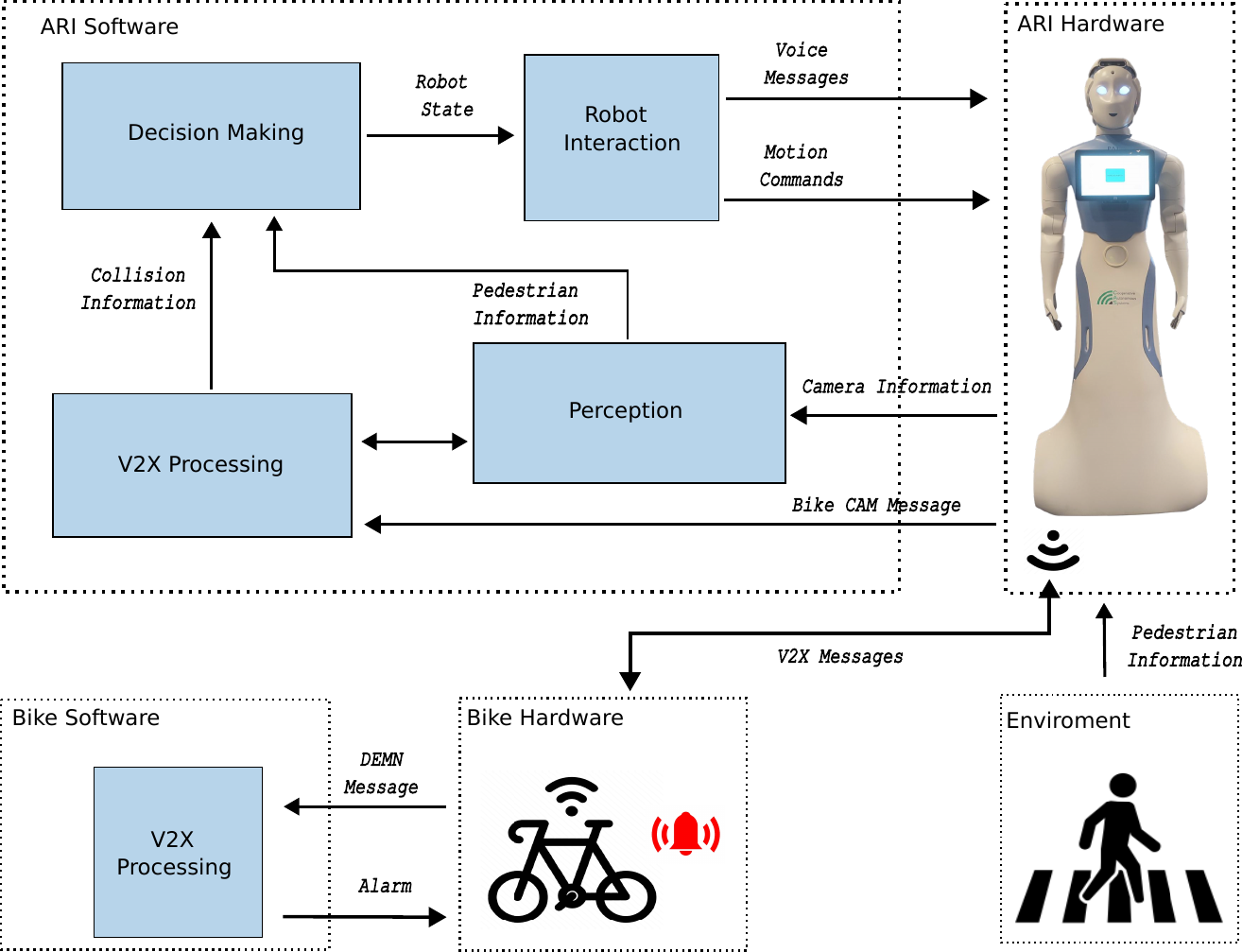}
\caption{Software architecture of our solution. The bike exchanges information with ARI robot through \ac{V2X} and the robot software use it together with pedestrian information to generate actions for the robot.}
\label{fig:architecture}
\vspace{-0.5cm}
\end{figure}

Social robot interaction in dynamic traffic environments requires a well-structured and robust system architecture to ensure reliable operation in both indoor and outdoor settings. The proposed ARI software framework is therefore organized into four main components, as illustrated in~Fig.~\ref{fig:architecture}, to enable safe, context-aware, and efficient interaction with pedestrians and connected vehicles. The perception module processes data from ARI’s onboard camera to detect pedestrians and infer their intentions in real time. All camera data are processed locally and are not stored, thereby addressing privacy considerations. The \ac{V2X} processing module manages the \ac{CAM} from connected road users, such as e-bikes, and extracts relevant information to assess potential risks. By combining perception and communication data, the system enhances its situational awareness and is able to anticipate possible interactions. The decision-making module utilizes a \ac{FSM} to analyze inputs from the perception and \ac{V2X} modules and determine appropriate system actions under varying environmental conditions. Finally, the interaction module executes the selected actions through motion commands and voice messages, enabling clear and effective communication with pedestrians.

In addition to the ARI robot, the system architecture also incorporates an intelligent e-bike equipped with its own \ac{V2X} processing module. This module is responsible for transmitting and interpreting \ac{V2X} messages, specifically \acp{CAM}, and \acp{DENM}. By integrating these components, the proposed architecture enables a seamless flow of information between the ARI robot, connected vehicles, and vulnerable road users. 

\subsection{Interaction principles}
\begin{figure}[!htb]
\centering
\includegraphics[width=.95\linewidth]{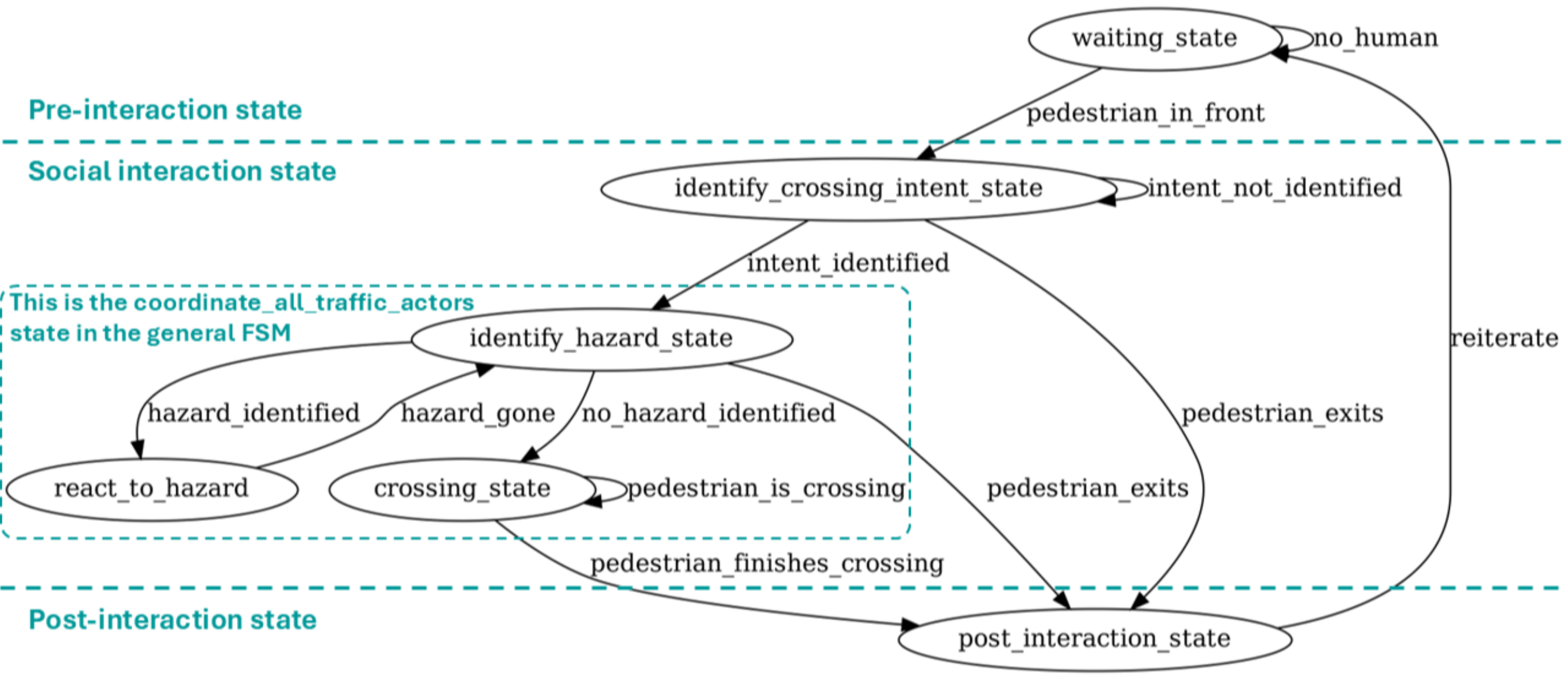}
\caption{State machine describing the interaction of the PoC.}
\label{fig:state_machine}
\vspace{-0.5cm}
\end{figure}

Fig.~\ref{fig:state_machine} illustrates the sequence of ARI's interaction with an approaching vehicle and a pedestrian aiming to cross the road in this traffic scenario and Fig.~\ref{fig:robot_actions} presents the corresponding gesture vocabulary. Initially, ARI resides in a \emph{waiting state} and perceives its environment for pedestrians using the RGB-D camera mounted on its forehead. When a pedestrian is detected, ARI transitions to the \emph{identify crossing intent state}, where it evaluates whether the pedestrian intends to cross in a 100 ms loop . In this PoC, a simplifying assumption is made: if the pedestrian is facing ARI, they are considered to have the intention to cross (this approach will be updated in future work with an actual intent identification). Based on this assumption, ARI proceeds to the \emph{identify hazard state} when the intent is true. In this state, ARI checks for hazards - namely, approaching vehicle towards the pedestrian. ARI gathers and computes \ac{CAM} from all surrounding vehicles within 400-meter range to identify potential threats. This is done by defining a zone of danger (\emph{ZoD}) close to the crossing area, computing times of entry and times of exit to the \emph{ZoD} for all detected vehicles and comparing these times, with a time threshold $T_{\text{safe}}$ .  
It is explained in detail in IV-D. Upon identification of the danger, ARI enters the \emph{react to hazard} in which it communicates with the pedestrian verbally (\emph{stop}) and non-verbally (\emph{hand gestures}) while transmitting a \ac{DENM} \emph{pedestrian in front} as a warning to the approaching vehicle. Additionally, the approaching vehicle can receive the ARI's \ac{CAM} for location awareness. ARI cycles between the \emph{react to hazard} and \emph{identify hazard state} until the hazard disappears (when the vehicle passes safely). When no hazards identified, ARI transitions to \emph{crossing state},  signaling the pedestrian to cross the road using verbal (\emph{you can cross}) and nonverbal (\emph{hand gesture}) cues. Finally, ARI enters the \emph{post interaction state} where the outcomes of the interaction are converted into relevant metrics and used to update decision-making parameters, allowing the system to refine itself. This iterative process ensures that each subsequent interaction results in improved performance before returning to the initial \emph{waiting state}, repeating the whole sequence.

\subsection{Implementation details}
\begin{figure}[t]
\centering
\includegraphics[width=.8\linewidth]{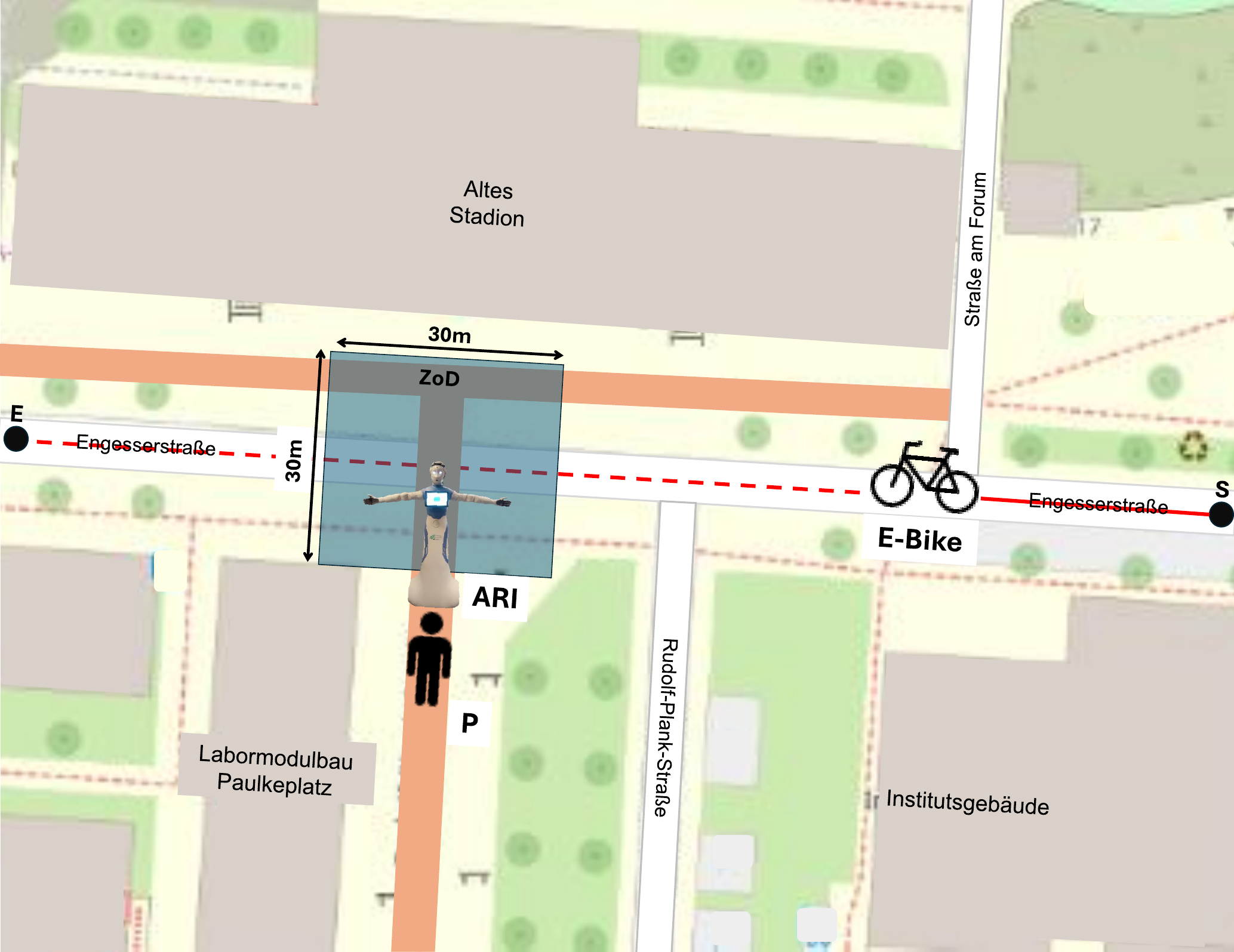}
\caption{Top view of the experimental setup. (P) denotes the pedestrian, while (S) and (E) indicate the starting and end points of the e-bike trajectory. In this scenario, the pedestrian approaches from the sidewalk to cross the main road, while the e-bike travels along the main road (red line indicating its predicted trajectory). ARI (the humanoid robot) intervenes by stopping the pedestrian and allowing crossing only after the e-bike has passed the Zone of Danger (ZoD). Map data © OpenStreetMap contributors.}
\label{fig:poc_top_view}
\vspace{-0.3cm}
\end{figure}

For our PoC implementation, we selected the same location at the KIT campus where the VR demo videos for pre-study described in Section~III~B were recorded. The site consists of a two-way single-lane road with a pedestrian priority area, where vehicle speeds are typically low, as shown in Fig. \ref{fig:poc_top_view}. As the area is managed by KIT, it enables controlled experimentation without affecting traffic. Furthermore, the relatively simple road layout reduces environmental variability and scenario complexity, which facilitates reproducible evaluation of the proposed system. 
As stated, we employ the humanoid robot ARI and an approaching e-bike for this PoC. We equipped both with \ac{V2X} OBUs, enabling wireless communication between them. Through this setup, both entities exchange \ac{CAM}, which provide essential state information--including station type, station identifier, latitude, longitude, speed, and heading--as well as \ac{DENM}, which convey event-driven information. 

For better evaluation of our PoC, we create \emph{ZoD}, which is computed respect to the robot pose. Here, ARI is positioned along the center of the zone's top border on the sidewalk to minimize disruption to road traffic, as illustrated in Fig.~\ref{fig:robot_state_a}. 

Let the position of ARI be represented by $ (x_{\text{ari}}, y_{\text{ari}})$. The ZoD is defined as a rectangular region centered relative to this reference position:
\begin{equation*}
\begin{aligned}
ZoD = \{(x,y)\mid \;& x_{\text{ari}} - L_{\text{left}} \le x \le x_{\text{ari}} + L_{\text{right}}, \\
                 & y_{\text{ari}} - W_{\text{down}} \le y \le y_{\text{ari}} + W_{\text{up}} \}
\end{aligned}
\end{equation*}
where $L_{\text{left}}$ and $L_{\text{right}}$ denote the horizontal extents of the region relative to ARI, and $W_{\text{down}}$ and $W_{\text{up}}$ represent the vertical extents. For the robot, it is better to project the ZoD respect to its pose which is also stored in global coordinates such that it can be used by other vehicles.  


At the start of the PoC, ARI remains in the \emph{waiting state} while observing the environment for any pedestrians using its forehead-mounted camera, as shown in Fig. \ref{fig:robot_state_a}. Upon detecting a pedestrian standing in front, ARI transitions from the \emph{waiting state} to \emph{identify crossing state}, as depicted in Fig. \ref{fig:robot_state_b}. By utilizing the skeleton model of the pedestrian and measuring its body orientation relative to the ARI's base frame within a range of 2m, the intention of crossing is predicted, prompting the state machine to transition to the \emph{identify hazard state}.

In the \emph{identify hazard state}, ARI detects  nearby \ac{V2X}-enabled vehicles by receiving and processing \ac{CAM} messages transmitted. Each OBU is associated with a unique station ID, allowing ARI to distinguish between participants. The approaching e-bike is identified through its corresponding station ID.

Let the initial detected position of the e-bike be $(x_{\text{eb}}(0), y_{\text{eb}}(0))$ with speed $v$ and heading angle $h$ (in radians). At each time step ($t = n \cdot 1 \, \text{s}$, where n is a non-negative integer), the predicted position of the e-bike at time $t$ is
\noindent
\begin{minipage}{0.49\columnwidth}
\begin{equation*}
x_{\text{eb}}(t) = x_{\text{eb}}(0) + v t \cos(h),
\end{equation*}
\end{minipage}
\hfill
\begin{minipage}{0.47\columnwidth}
\begin{equation*}
y_{\text{eb}}(t) = y_{\text{eb}}(0) + v t \sin(h).
\end{equation*}
\end{minipage}
\newline

ARI determines the entry time $t_{\text{entry}}$ and exit time $t_{\text{exit}}$ at which the predicted trajectory intersects the ZoD:

\begin{equation*}
(x_{\text{eb}}(t), y_{\text{eb}}(t)) \in ZoD.
\end{equation*}

Using the equations for the e-bike’s trajectory, solve for t
satisfying the boundaries:


\noindent
\begin{minipage}{0.49\columnwidth}
\begin{equation*}
t_{x}^{\text{enter}} =
\frac{x_{\text{ari}} - L_{\text{left}} - x_{\text{eb}}(0)}
{v\cos(h)}
\end{equation*}
\end{minipage}
\hfill
\begin{minipage}{0.47\columnwidth}
\begin{equation*}
t_{x}^{\text{exit}} =
\frac{x_{\text{ari}} + L_{\text{right}} - x_{\text{eb}}(0)}
{v\cos(h)}
\end{equation*}
\end{minipage}



\noindent
\begin{minipage}{0.49\columnwidth}
\begin{equation*}
t_{y}^{\text{enter}} =
\frac{y_{\text{ari}} - W_{\text{down}} - y_{\text{eb}}(0)}
{v\sin(h)}
\end{equation*}
\end{minipage}
\hfill
\begin{minipage}{0.47\columnwidth}
\begin{equation*}
t_{y}^{\text{exit}} =
\frac{y_{\text{ari}} + W_{\text{up}} - y_{\text{eb}}(0)}
{v\sin(h)}
\end{equation*}
\end{minipage}
\newline


The time interval during which the e-bike is predicted to be inside the ZoD is therefore

\begin{minipage}{0.49\columnwidth}
\begin{equation*}
t_{\text{entry}} = \max(t_{x}^{\text{enter}}, t_{y}^{\text{enter}})
\end{equation*}
\end{minipage}
\hfill
\begin{minipage}{0.45\columnwidth}
\begin{equation*}
t_{\text{exit}} = \min(t_{x}^{\text{exit}}, t_{y}^{\text{exit}})
\end{equation*}
\end{minipage}
\newline

\begin{figure}[t]
    \centering
    \begin{subfigure}{0.2\textwidth}
        \centering
        \fbox{\includegraphics[width=0.8\textwidth]{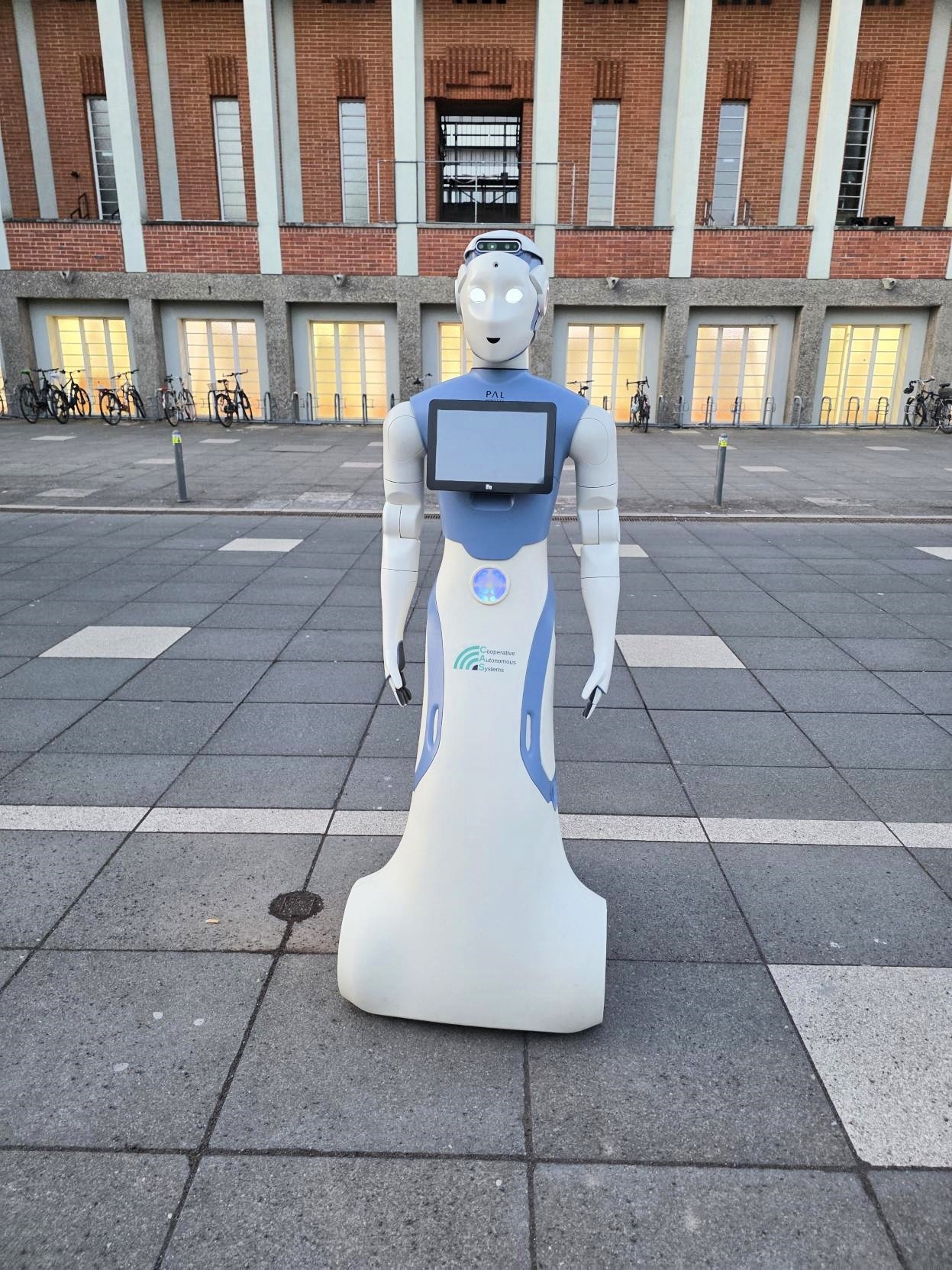}}
        \caption{}
        \label{fig:robot_state_a}
    \end{subfigure}
    \begin{subfigure}{0.2\textwidth}
        \centering
        \fbox{\includegraphics[width=0.8\textwidth, height=1.06\textwidth]{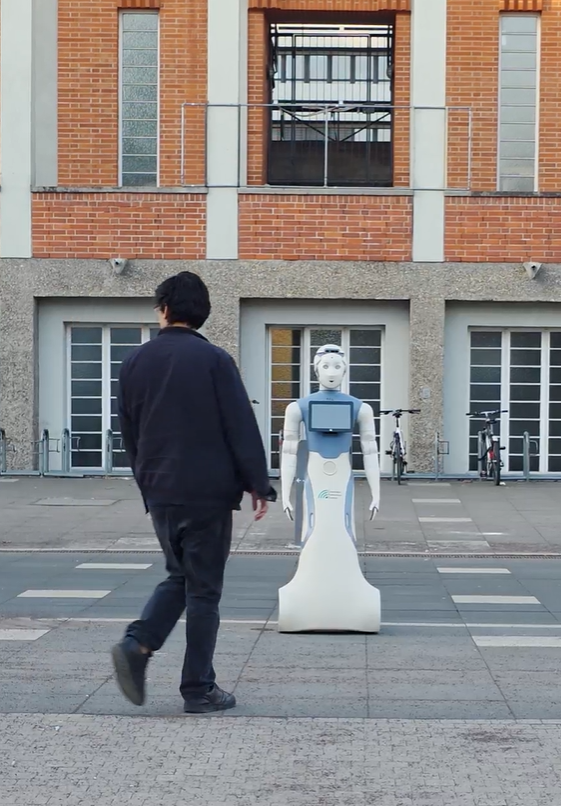}}
        \caption{}
        \label{fig:robot_state_b}
    \end{subfigure}
    \\
    \begin{subfigure}{0.2\textwidth}
        \centering
        \fbox{\includegraphics[width=0.8\textwidth]{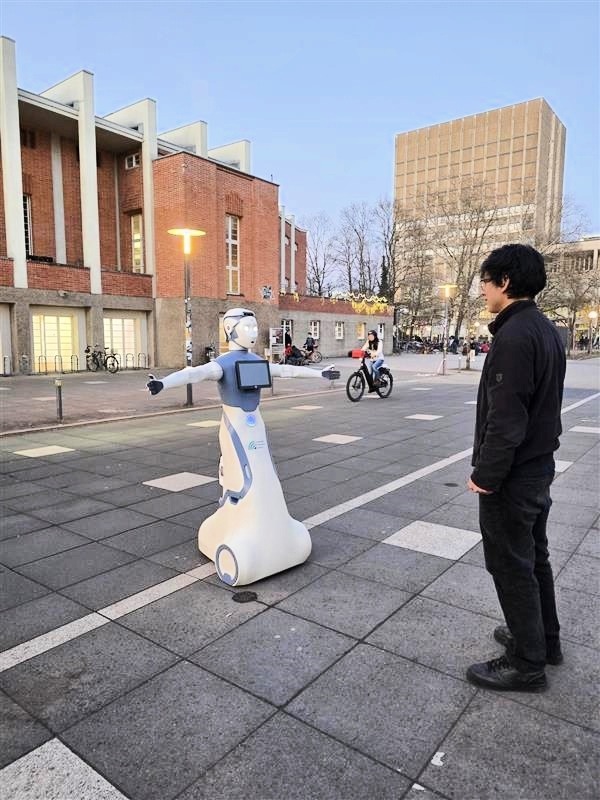}}
        \caption{}
        \label{fig:robot_state_c}
    \end{subfigure}
    \begin{subfigure}{0.2\textwidth}
        \centering
        \fbox{\includegraphics[width=0.8\textwidth]{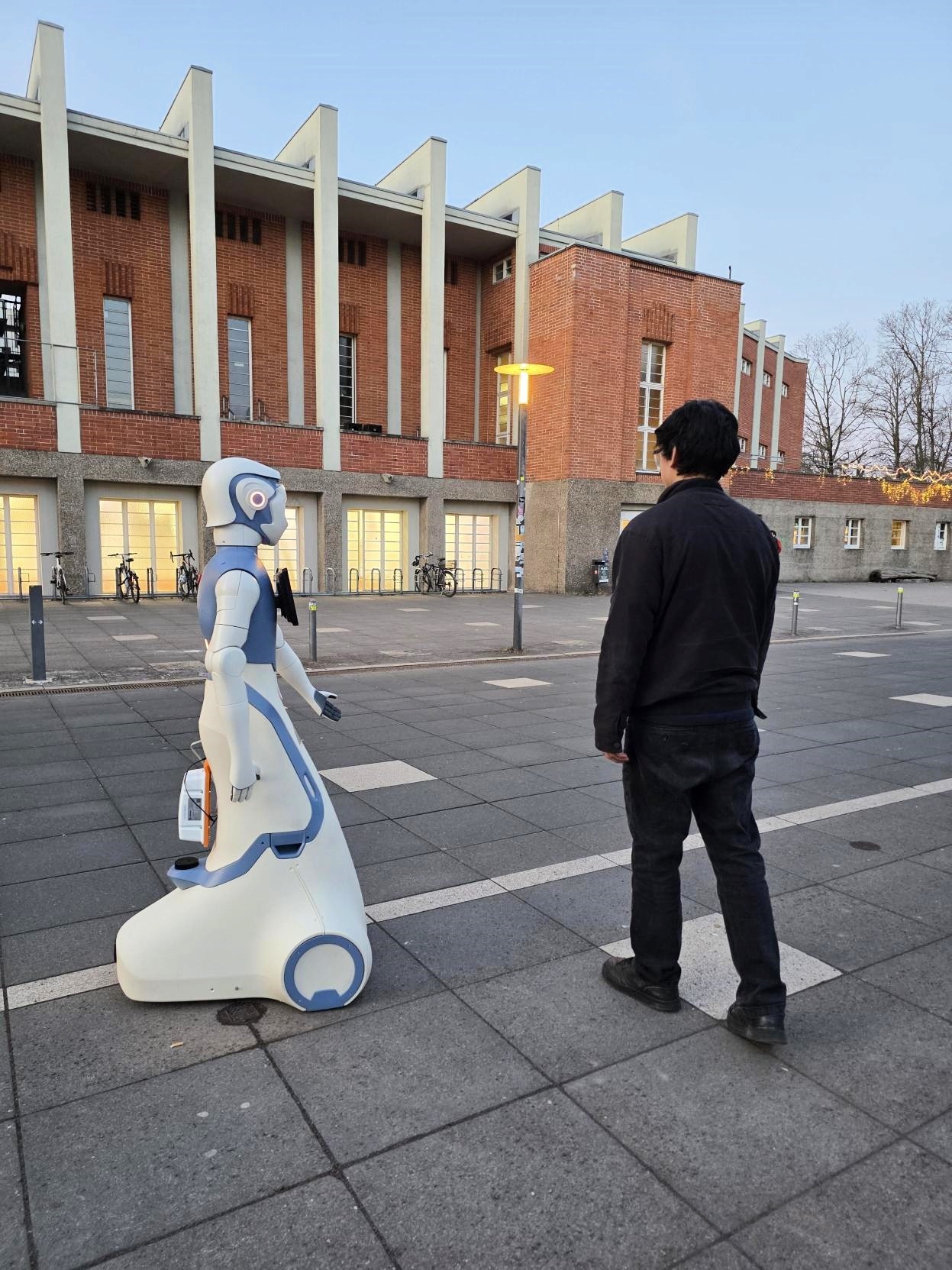}}
        \caption{}
        \label{fig:robot_state_d}
    \end{subfigure}
    \caption{Illustration of the \ac{FSM} of ARI. (a) \emph{Waiting state} – ARI passively observes its environment. (b) \emph{Identify crossing state} – ARI assesses the pedestrian’s intent to cross. (c) \emph{Identify hazard state} – An e-bike (seen with a red bag in the background) enters the ZoD, prompting ARI to signal a ``stop” and transmit a DENM. (d) \emph{Crossing state} – ARI permits pedestrian crossing with a hand gesture and terminates the transmitted DENM.}
    \vspace{-0.5cm}
    \label{fig:robot_actions}
\end{figure}

\begin{figure}[h]
    \centering
    \begin{subfigure}{0.2\textwidth}
        \centering
        \fbox{\includegraphics[width=0.8\textwidth]{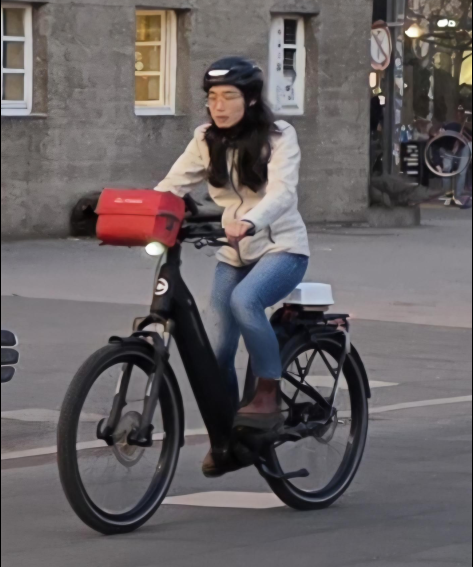}}
        \caption{}
        \label{fig:robot_state_b}
    \end{subfigure}
    \begin{subfigure}{0.2\textwidth}
        \centering
        \fbox{\includegraphics[width=0.8\textwidth]{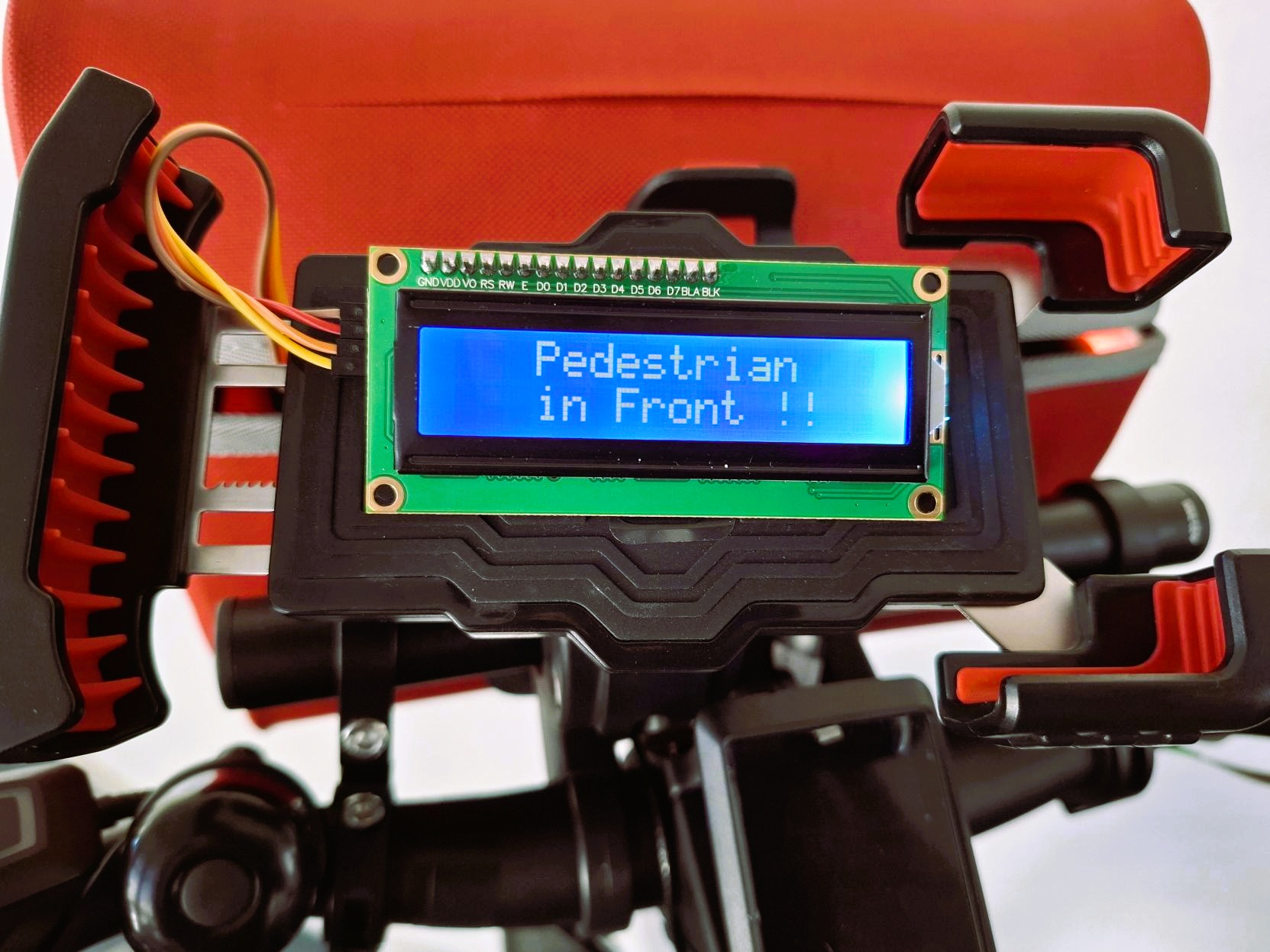}}
        \caption{}
        \label{fig:robot_state_a}
    \end{subfigure}
    \caption{Illustration of the e-bike used in the PoC scenario. (a) A rider riding the e-bike inside the ZoD. (b) The LCD display installed on the e-bike presenting a “Pedestrian In Front" warning to the user based on the \ac{DENM} transmitted by ARI.}
    \label{fig:bike_display}
    \vspace{-0.5cm}
\end{figure}

If $
t_{\text{entry}} > t_{\text{exit}},
$ the predicted trajectory does not intersect the ZoD. In this case, ARI determines that the crossing maneuver is safe and signals the pedestrian to proceed using a hand gesture and the auditory cue ``You can cross'' (Fig.~\ref{fig:robot_state_d}). \\
Conversely, If the trajectory intersects the ZoD, ($t_{\text{entry}} \le t_{\text{exit}}$) and the predicted entry time satisfies $t_{\text{entry}} < T_{\text{safe}}$, where $T_{\text{safe}} = 5\,\text{s}$.  ARI prevents the pedestrian from crossing by displaying a stop gesture and issuing a verbal warning, as illustrated in Fig. \ref{fig:robot_state_c}. Additionally, ARI transmits a \ac{DENM} with the cause code \emph{HumanPresenceOnTheRoad}. This message is received by the e-bike OBU and displayed to the rider as a warning (Fig.~\ref{fig:bike_display}). The notification is advisory and does not influence vehicle control.

Once the e-bike exits the ZoD, as determined by $t_{\text{entry}} \geq t_{\text{exit}}$, ARI transitions to \emph{crossing state} and permits the pedestrian to cross and terminates the transmitted DENM. Notably, when the e-bike is within the ZoD, $t_{\text{entry}}$ will be negative and $t_{\text{exit}}$ will be positive, signifying an active danger situation. Upon the successful crossing of the pedestrian, ARI transitions to the final \emph{post-interaction state}, wherein the outcomes of the interaction are recorded and utilized to update decision-making parameters, thereby refining the system. Subsequently, ARI reverts to the initial \emph{waiting state}. 

\subsection{Lessons learned}
To validate the PoC, eight trials were conducted in which the e-bike traversed a fixed route from the crossing area, passing through the ZoD and exiting on the far side, while ARI managed the pedestrian crossing. The ZoD was defined as a 30×30\,m centered at the robot's position. Three quantitative metrics were extracted from the received \ac{CAM} data to provide context on the kinematic severity of each trial:

\begin{itemize}
    \item \textit{Decision Distance (DD)}: the distance between the e-bike and the ZoD boundary at the moment ARI issued the stop intervention.
    \item \textit{ZoD Occupancy Duration}: the time the e-bike spent within the ZoD, which also corresponds to the pedestrian's waiting time, as ARI holds the pedestrian until the e-bike exits the zone.
    \item \textit{Bike Speed at ZoD Entry}: the speed of the e-bike at the moment it entered the ZoD boundary.
\end{itemize}

\begin{table}[h]
\vspace{-0.2cm}
\centering
\caption{Quantitative Metrics Across Eight Trials. 
Values reported as mean $\pm$ standard deviation ($N=8$).}
\label{tab:metrics}
\begin{tabular}{lccc}
\toprule
\textbf{Trial} & \textbf{DD (m)} & \textbf{ZoD Occ. (s)} & \textbf{Speed at Entry (m/s)} \\
\midrule
1  & 17.88 & 7.61  & 4.50 \\
2  & 19.15 & 10.53 & 3.69 \\
3  & 16.43 & 8.79  & 2.43 \\
4  & 15.24 & 10.04 & 3.19 \\
5  & 14.67 & 8.99  & 3.08 \\
6  & 15.10 & 9.02  & 3.40 \\
7  & 14.58 & 9.20  & 3.08 \\
8  & 16.02 & 7.80  & 3.64 \\
\midrule
\textbf{Mean $\pm$ Std} & $16.13 \pm 1.63$ & $9.00 \pm 0.99$ & $3.38 \pm 0.60$ \\
\bottomrule
\end{tabular}
\vspace{-0.5cm}
\end{table}

The results across eight trials are summarized in Table~\ref{tab:metrics}. ARI consistently triggered the stop intervention while the e-bike was still outside the ZoD, with a mean DD of $16.13 \pm 1.63\,\mathrm{m}$, confirming reliable early detection enabled by \ac{V2X} communication. The mean ZoD occupancy duration was $9.00 \pm 0.99\,\mathrm{s}$, representing a predictable pedestrian waiting period. The mean bike speed at ZoD entry was $3.38 \pm 0.60\,\mathrm{m/s}$, reflecting naturalistic riding conditions across trials. The low variance observed across all three metrics demonstrates the repeatability of the system's behavior, and confirms that \ac{V2X}-assisted early detection allows ARI to intervene with sufficient margin before any potential conflict arises.

\section{Technical Enablers}

The pedestrian-crossing assistant introduced in the previous section can be generalized into a broader system concept in which a \ac{V2X}-enabled social robot operates as an intersection coordination unit for mixed traffic~\cite{Seitz2022}. In this role, the ROBOPOL is not merely an interface to pedestrians, but an active coordination entity that supports safe and efficient cooperative automated driving by combining local sensing, connected information exchange, traffic reasoning, and multimodal interaction, as illustrated in Fig.~\ref{V2X4Robot}. To fulfill this role, the robot must provide three core capabilities: (1) situational awareness of all relevant traffic participants and environmental conditions, including partially occluded actors; (2) intention inference and conflict detection based on observed states, communicated information, and traffic context resulting in a decision-making under uncertainty to derive safe, efficient, and fair coordination actions; and (3) multimodal communication with both connected vehicles and VRUs through machine-readable and human-understandable interfaces. 

These requirements define three tightly coupled technical enablers discussed in the following: perception and situational awareness, planning and formal reasoning, vehicular communication and social interaction. Together, these enablers form the system backbone of robot-assisted intersection management. Each enabler is presented in terms of the specific system requirement it fulfills and its role within the overall coordination loop.

\subsection{Enabler 1 -- Perception and situational awareness}


The perception enabler supports the robot’s requirement for real-time situational awareness by constructing a consistent and occlusion-resilient representation of all relevant traffic actors. This is achieved through the integration of onboard sensing and \ac{V2X}-enabled collective perception, enabling the robot to maintain an up-to-date and shared view of the intersection environment. In the scenario in Fig. \ref{fig:poc_top_view}, the traffic actors are the vehicles perceived with V2X and the pedestrians tracked with the robot head camera.

\textbf{Singleton Perception.}
At the local level, the robot relies on onboard sensors to detect, track, and interpret the behavior of nearby traffic participants~\cite{siegwart2011,Camara2020a}. The key objective is not the specific sensing modality, but the ability to maintain a dynamic representation of the environment that includes the position, velocity, and class of each relevant actor, as well as higher-level attributes such as pedestrian pose, gaze direction, and gestures.

This perception pipeline consists of detection, tracking, and prediction stages. Detection identifies relevant objects in the scene, tracking maintains their state over time, and prediction estimates their short-term evolution. In the case of VRUs, prediction is particularly challenging due to the variability of human behavior, requiring models that account for contextual and behavioral cues. The output of this pipeline is a structured scene representation that captures the current traffic situation and serves as input to subsequent reasoning and planning modules. Different sensing modalities, such as cameras, radar, and lidar, offer complementary trade-offs in range, robustness, and resolution, motivating their combined use in practice.

\textbf{Collective Perception.}
To overcome the inherent limitations of local sensing, the robot leverages collective perception through \ac{V2X} communication. By receiving information from connected vehicles and roadside units~\cite{Markgraf2023}, the robot extends its situational awareness beyond its direct line of sight, which is essential in intersection scenarios characterized by frequent occlusions and dynamically evolving traffic conditions.

\acp{CPM} provide abstract representations of detected objects, including their position, motion, and classification. These messages enable interoperability across heterogeneous platforms and allow the robot to integrate externally observed objects into its local scene representation. Through data fusion, the robot combines onboard sensor data with information received via \acp{CPM}, resulting in a more accurate and comprehensive understanding of the environment.

This shared perception capability is particularly important for early detection of potential conflicts, such as vehicles or pedestrians approaching from occluded areas. By incorporating \ac{CPM} data, the robot can anticipate such situations and support timely coordination decisions. The standardization of \acp{CPM} by \ac{ETSI}~\cite{ETSI_TS_103_324} ensures compatibility across systems and facilitates their deployment in cooperative automated driving scenarios~\cite{Delooz2023}.

Despite these advantages, collective perception introduces challenges related to communication latency, data volume, and reliability. Ensuring timely transmission and consistent fusion of distributed observations is therefore critical to maintain the relevance and accuracy of the shared environment model.

The resulting fused scene representation constitutes the input to the planning and reasoning modules, enabling the robot to infer intentions and derive coordinated actions based on a comprehensive and up-to-date view of the intersection.

\subsection{Enabler 2 -- Planning and formal reasoning}

The planning enabler supports intention inference and decision-making under uncertainty by combining stochastic traffic modeling with formal reasoning to derive safe, efficient, and fair coordination actions. Based on the fused scene representation obtained from perception and collective perception, the robot continuously evaluates the current traffic situation and determines coordinated actions for all relevant actors at the intersection~\cite{fiems2006throughput}.

\textbf{Stochastic Modeling.}
Intersection traffic is inherently stochastic due to uncertainties in arrival processes, motion dynamics, communication delays, and human behavior. To account for this variability, stochastic models are used to represent the evolution of traffic flows and to evaluate the impact of different coordination strategies. In the context of the robot-assisted intersection, these models support the estimation of delays~\cite{Boon2019}, queue evolution, and potential conflicts under varying traffic conditions~\cite{Timmerman2024,WOS:001307654100001,macedo2017application,WOS:000662126600001}.

Key sources of uncertainty include the timing and type of arriving road users, their route choices, variability in maneuver execution, perception noise, and communication imperfections in \ac{V2X} systems. By incorporating these factors, the robot can assess the likelihood of future traffic states and evaluate candidate coordination actions accordingly.

From a decision-making perspective, the robot must dynamically allocate intersection access among competing traffic participants. This can be formulated as a stochastic scheduling problem, where each road user corresponds to a job requiring access to shared spatial resources. Approaches such as Markov decision processes and reinforcement learning enable the derivation of adaptive control policies, while queueing-based models provide tractable approximations that support real-time operation~\cite{Terekhov2014,haijema2017dynamic,WOS:000815825900008,WOS:000759387900001}.

Rather than relying on a single modeling paradigm, the robot benefits from combining these approaches: stochastic models capture system dynamics and uncertainty, while simplified scheduling abstractions enable efficient computation of coordination decisions in real time.

\textbf{Formal Reasoning.}
While stochastic modeling enables optimization under uncertainty, formal reasoning ensures that derived actions comply with traffic regulations~\cite{CE16,BTK07,RakowSchwammberger2023} and safety constraints~\cite{Rand16}. To this end, the robot translates the perceived traffic situation into a machine-readable representation that supports logical reasoning about spatial relations, temporal evolution, and rule compliance.

In particular, traffic logics, such as Urban Multi-lane Spatial Logic (UMLSL)~\cite{Sch18-TCS}, provide a formal framework to describe traffic configurations and reason about safe maneuvers. By combining such logical representations with agent models and formalized traffic rules~\cite{SA21}, the robot can evaluate whether candidate actions violate safety constraints or regulatory requirements. 
Formal reasoning is especially critical in conflict situations, where competing objectives or rules must be balanced~\cite{SchwammbergerIV25}. In such cases, the robot must resolve trade-offs between safety, efficiency, and fairness. Approaches based on game-theoretic reasoning~\cite{traffic-games24} allow the robot to analyze interactions among multiple agents and select coordination strategies that satisfy system-level objectives while respecting individual constraints.

\textbf{Integrated Decision-Making.}
The planning process integrates stochastic modeling and formal reasoning within a unified decision loop. Stochastic models generate and evaluate candidate coordination actions under uncertainty, while formal reasoning filters and constrains these actions to ensure safety and rule compliance. The resulting decisions include, for example, granting or delaying access to the intersection, assigning priorities among road users, or generating speed advisories for connected vehicles.

These decisions are continuously updated as new perception and \ac{V2X} information becomes available, enabling adaptive and context-aware intersection management. The outputs of the planning module are then passed to the interaction and communication enabler, where they are translated into actionable instructions for both connected vehicles and VRUs.

\subsection{Enabler 3 - V2X communication and social interaction}
The interaction enabler supports the execution of coordination decisions by enabling the robot to communicate actions to connected vehicles via \ac{V2X} and to VRUs through human-understandable multimodal interfaces. Acting as the intersection coordination unit, the robot translates planning outputs into actionable instructions tailored to heterogeneous traffic participants.

\textbf{Human-Robot Interaction.}
For VRUs, the robot must convey coordination decisions in a clear, intuitive, and unambiguous manner. The primary communication modality is visual signaling through body gestures, which can indicate actions such as stopping, yielding, or proceeding. These gestures can be complemented by visual displays to reinforce intent and improve interpretability in complex scenarios.

The design of such interaction mechanisms must ensure rapid comprehension and minimize ambiguity, particularly in safety-critical situations. Compared to acoustic communication, visual modalities are more robust in noisy urban environments and less prone to interference from multiple sources. The robot’s physical embodiment provides an advantage over static interfaces, as motion and spatial positioning can be used to direct attention and emphasize instructions.

A key challenge lies in achieving an appropriate balance of authority and trust. The robot must be perceived as a reliable coordination entity whose instructions are followed, while avoiding both under-trust, which reduces effectiveness, and over-trust, which may lead to unsafe behavior if system errors occur. Therefore, interaction design must align with human expectations in traffic contexts and remain consistent with established traffic conventions.

\textbf{Vehicle-Robot Interaction via \ac{V2X}.}
For connected vehicles, interaction is realized through standardized \ac{V2X} communication. Based on the planning outputs, the robot disseminates coordination information such as priority assignments, speed advisories, or hazard notifications. Existing message sets, \ac{CAM} and \ac{DENM}, support the exchange of status and event-driven information.

In addition, collective perception information is exchanged via \acp{CPM}, enabling vehicles to benefit from the robot’s extended situational awareness. However, due to the robot’s active coordination role, existing message sets are insufficient to fully represent its state and intent. To address this, a dedicated Robotic Awareness Service (RAS) can be introduced, allowing the robot to broadcast its planned actions and coordination decisions to surrounding vehicles.

RAS messages include dynamic information about the robot’s current state, intended coordination actions, and relevant interaction context. Their transmission characteristics must balance responsiveness and communication efficiency, with update rates aligned between those of vehicle-centric (e.g., \ac{CAM}) and vulnerable-road-user-centric messaging schemes.

The robot is expected to operate across multiple \ac{V2X} technologies, including IEEE 802.11p, LTE-\ac{V2X}, and 5G NR-\ac{V2X}, enabling interoperability in heterogeneous communication environments. In decentralized communication modes, the robot can directly exchange information with vehicles and infrastructure, ensuring low-latency dissemination of coordination decisions.

\textbf{Integration within the Coordination Loop.}
The interaction enabler closes the perception-planning-action loop by translating computed coordination strategies into executable signals. Decisions derived by the planning module are continuously communicated to traffic participants through both \ac{V2X} messages and human-interpretable signals, ensuring that all actors receive consistent and synchronized instructions.

This dual-interface capability is essential for mixed traffic environments, where connected and non-connected participants coexist. By combining machine-readable communication with intuitive human interaction, the robot enables coordinated behavior across heterogeneous agents and supports safe and efficient intersection management.

\subsection{System integration: Architectural synthesis and operational flow}
\label{subsec:integration}

\begin{figure}[t]
\centering
\includegraphics[width=.95\linewidth]{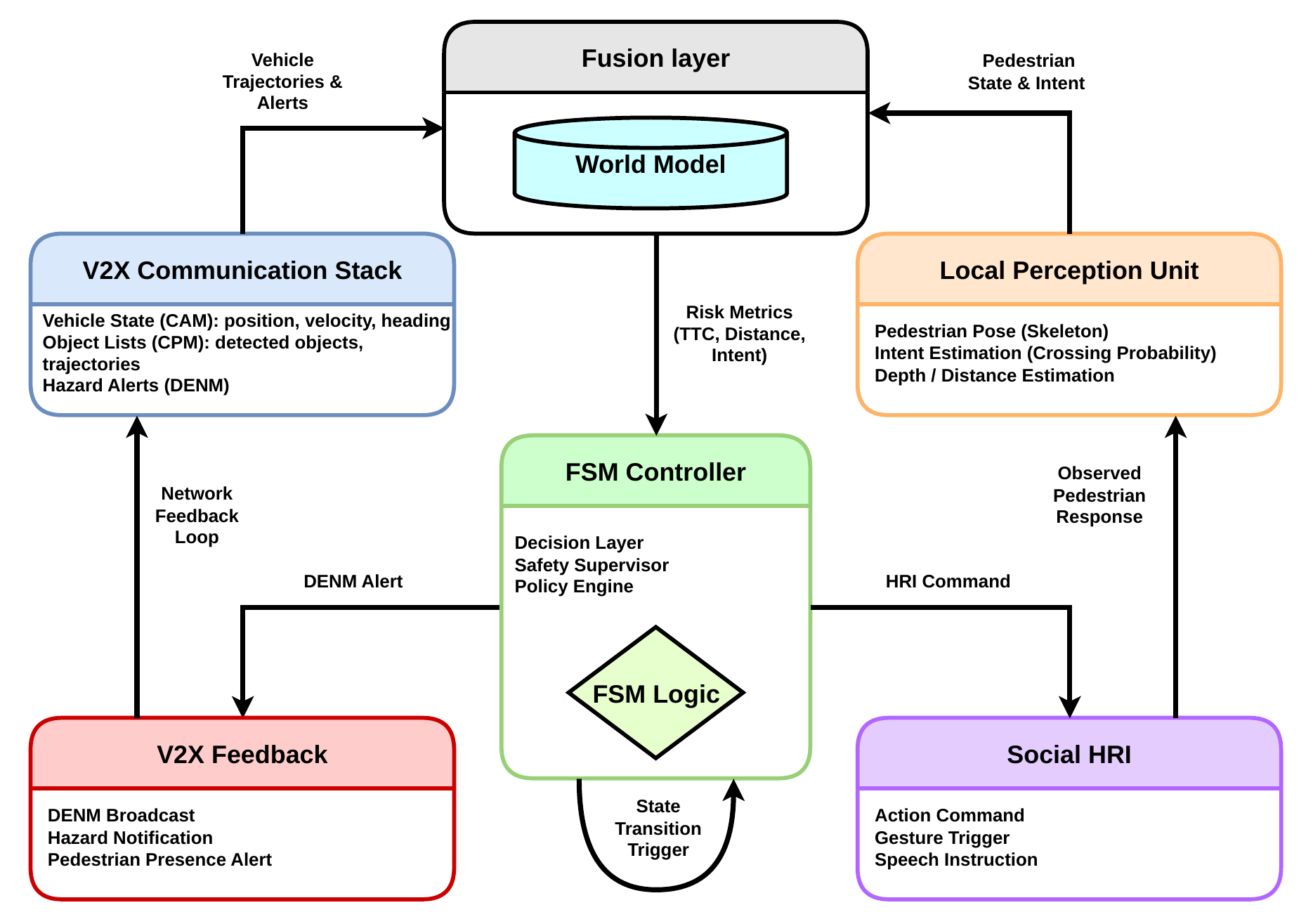}
\caption{Architectural synthesis and operational flow.}
\label{fig:enablers}
\vspace{-0.5cm}
\end{figure}

The efficacy of the ROBOPOL framework is predicated not upon the isolated performance of its components, but rather on the seamless orchestration of the technical enablers detailed in previous sections. To transition from discrete technical domains to a functional system-level concept, these enablers are integrated into a unified, closed-loop control architecture that transforms the robot into an active intersection coordination unit (see Fig. \ref{fig:enablers}).

The system functions as a multi-modal coordination hub, where information is acquired, interpreted, and translated through a continuous processing pipeline:
\begin{itemize}
    \item Perception and fusion module forms the foundation of the system’s dynamic scene representation. The robot’s local sensing suite, using skeleton tracking and pose estimation, provides ground truth on \ac{VRU} behavior. This ego-centric view is augmented by the \ac{V2X} stack, which ingests \acp{CPM}, \acp{CAM}, and \acp{DENM}. By aligning global \ac{V2X} data with the robot’s local frame, the system extends awareness beyond line-of-sight.

    \item Cognitive planning and orchestration uses the fused scene model as input to the coordination logic. As described in Section \ref{sec:fsm_logic}, behavior is governed by an \ac{FSM}, which processes environmental inputs to infer intent and evaluate coordination strategies. This stage derives key spatio-temporal parameters, such as priority and dynamic ZoD boundaries, triggering state transitions.

    \item Cross-layer interaction module executes \ac{FSM} decisions via dual-domain translation. It generates machine-readable \ac{V2X} messages for connected agents and triggers HRI modalities for humans. This converts digital risk assessments into intuitive physical signals, such as gestures and acoustic cues, bridging digital and physical traffic layers.ic warnings, ensuring that the robot serves as a tangible bridge between the digital and physical traffic layers.
\end{itemize}

The ROBOPOL system operates as an iterative perception-planning-interaction loop. At each cycle, the internal world model is updated with high-frequency sensor data and V2X telemetry, triggering an immediate re-evaluation of the traffic state. This iterative approach enables adaptive behavior in highly dynamic environments, ensuring that coordination decisions (such as a ``Stop'' gesture or a clearance signal) remain consistent with the most recent available information. The stability of this loop is maintained by balancing sensing update rates against communication latency, ensuring that the robot’s physical interventions are both timely and predictable for all road users~\cite{RotterDENM}. 

\section{Conclusion \& Outlook}
In this work, we take a broader perspective on the requirements to successfully deploy social robots equipped with vehicular communication technology in the traffic system. Robots integrated into C-ITS can bring new capabilities to enhance safety and efficiency in various scenarios. In robot-human interaction for pedestrian crossing, the robot can actively manage the interaction between pedestrians and vehicles, including e-bikes, by monitoring the speed and proximity of approaching vehicles and signaling them to slow down or by physically blocking the pedestrian if vehicles are moving too fast.  We identify four key enablers (advanced perception, \ac{V2X}, HRI and formalization methods) and give a short literature review of these key enablers necessary for such a system. We demonstrate a PoC of the world-first social robot solution for traffic orchestration using \ac{V2X}. For our PoC, we integrate the first three key enablers to demonstrate the technical feasibility of our approach. While most of the components of our presented approach need refinement, our presented PoC provides and important and necessary step to develop and deploy such a system. Next steps include extending the approach with formal methods allowing to analyze traffic situations, improving the perception system to make more accurate predictions of \acp{VRU}' behavior and improve understandability of the used gesture set. 

In the future, robots could assist autonomous vehicles to find free parking spots, the robot can scout for available spaces and signal their locations to nearby vehicles, reducing the time it takes to search for parking. For ensuring pedestrian pathway safety, the robot can detect e-bikes encroaching on pedestrian areas and take action to slow down or block their movement, preventing potential collisions. As a road safety assistant, the robot can coordinate vehicle maneuvers in complex traffic scenarios by gathering real-time data from the collective perception system and issuing coordinated instructions to optimize traffic flow and prevent accidents.

Nevertheless, a single robot is insufficient to effectively coordinate and manage complex traffic scenarios involving multiple road users across multi-lane environments\cite{FloresComeca2025SocialRF, Comeca2025RobotsFS}. This limitation highlights the necessity of extending the framework toward multi-robot coordination within the C-ITS ecosystem for traffic management. Additionally, systematic design of the \ac{V2X} protocols supporting inter-robot communications is required to enable scalable, reliable, and coordinated operation between different road users and robots. Furthermore, the scalability of the proposed PoC must be rigorously evaluated in large-scale multi-agent scenarios. Future investigations will also focus on identifying and analyzing failure modes, edge cases, and system-level limitations to ensure robustness and safe operation under diverse traffic conditions.

The ROBOPOL operates under good visibility, low traffic complexity, and available \ac{V2X} connectivity. More complex scenarios (e.g., crowds, non-connected vehicles, severe occlusions, adverse weather, or degraded communications) are outside the operational design domain. The robot provides advisory guidance only and does not replace pedestrian judgment. Adherence to the design principles for functional safety defined in ISO 26262, as well as the safety of the intended functionality ISO 21448, will be crucial for future adoption~\cite{SOTIF}. Robotics perception components that incorporate artificial intelligence are expected to undergo safety testing processes similar to those foreseen for autonomous vehicles.

Last but not least, a new regulatory framework will be required for the deployment of moderating traffic robots in real-world environments. These robots can be considered a form of mobile infrastructure, as they perform coordination functions similar to traffic lights while simultaneously possessing mobility capabilities comparable to vehicles. Therefore, the legal framework should incorporate best practices from traffic light controller safety certification as well as lessons learned from mobile robots already deployed in public spaces, such as street delivery robots.

\bibliographystyle{IEEEtran}
\bibliography{ref}


\newpage
\begin{IEEEbiography}[{\includegraphics[width=1in,height=1.25in,clip,keepaspectratio]{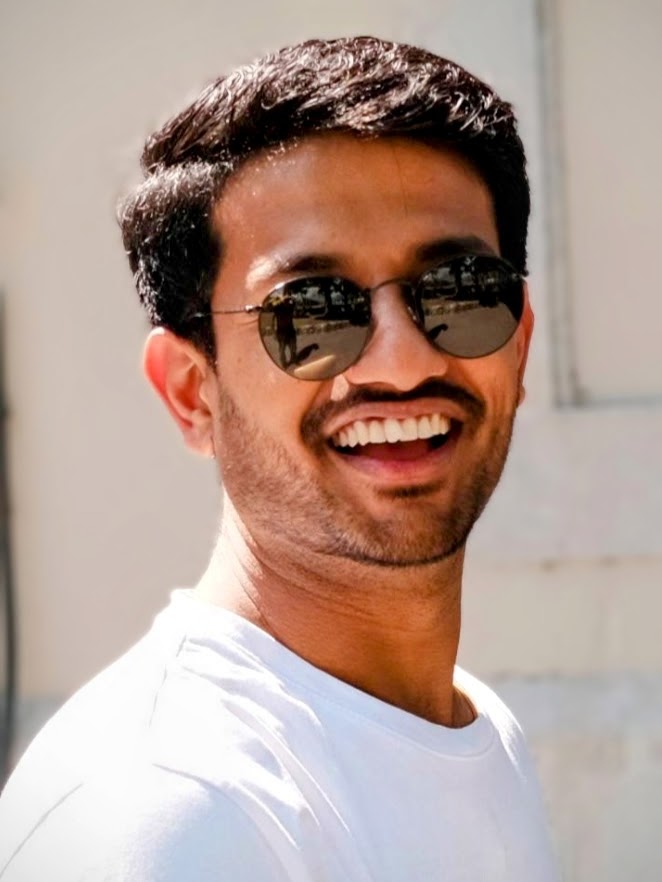}}]{John Pravin Arockiasamy}
is a doctoral candidate at the Karlsruhe Institute of Technology (KIT). He earned his Master’s degree in Information Technology from the University of Stuttgart in 2024. His research focuses on improving safety for vulnerable road users, and enhancing communication between vehicles and robots using \ac{V2X} technology. This work contributes to the development of cooperative autonomous systems that work effectively in real-world environments.
\end{IEEEbiography}

\begin{IEEEbiography}
[{\includegraphics[width=1in,height=1.25in,clip,keepaspectratio]{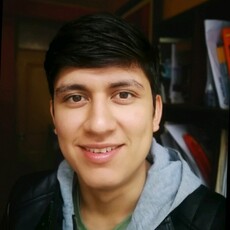}}]{Andy Flores Comeca} is a PhD Student at Karlsruhe Institut of Technology (KIT), Germany. He received his Bsc. degree in Electronics Engineering from Pontifical Catholic University of Peru (PUCP) and his M.Sc. degree Mechatronics from Technische Universität Ilmenau (TU-Ilmenau), Germany. His research fields are related to autonomous robots, human robot interaction and \ac{V2X} communication. 
\end{IEEEbiography}

\begin{IEEEbiography}
[{\includegraphics[width=1in,height=1.25in,clip,keepaspectratio]{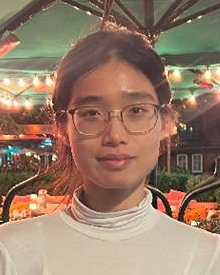}}]{Victoria (Ya-Ting) Yang} is a PhD student in the Socially Assistive Robotics with Artificial Intelligence (SARAI) group at KIT. She holds Bachelor’s and Master’s degrees from the University of Waterloo, Canada. Her research focuses on enhancing robots’ social interaction capabilities in dynamic, unstructured environments.
\end{IEEEbiography}

\begin{IEEEbiography}[{\includegraphics[width=1in,height=1.25in,clip,keepaspectratio]{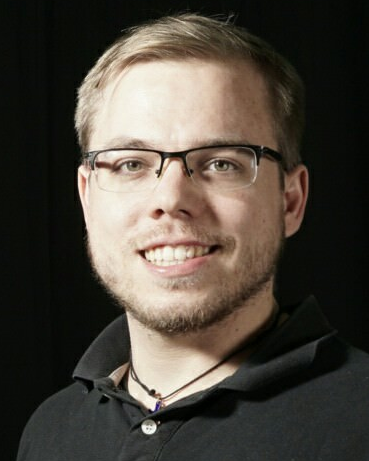}}]{Manuel Bied}
is a postdoctoral researcher at the Karlsruhe Institute of Technology (KIT), Germany. He received his Ph.D. degree in robotics from Sorbonne University, France and his Master's and Bachelor's degree in Electrical Engineering and Information Technology from Technical University Darmstadt, Germany.  His research interests include human-robot interaction, \ac{V2X}, machine learning, and collective perception. In his work, he is focusing on the use of robots in traffic with the aim of increasing road safety.
\end{IEEEbiography}

\begin{IEEEbiography}[{\includegraphics[width=1in,height=1.25in,clip,keepaspectratio]{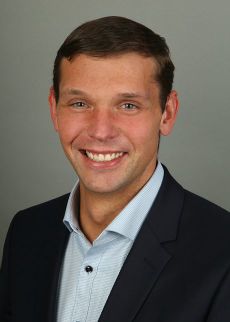}}]{Maximilian Schrapel}
is a postdoctoral researcher at the Karlsruhe Institute of Technology (KIT), Germany. He researches at the intersection of cooperative autonomous systems and vulnerable road users in intelligent, interconnected environments. His research is part of the Real World Lab Initiative, where novel \ac{V2X} demonstrators are being developed to conduct experiments in realistic scenarios. He specializes in Cooperative Intelligent Transport Systems and connected e-bikes.
\end{IEEEbiography}

\begin{IEEEbiography}[{\includegraphics[width=1in,height=1.25in,clip,keepaspectratio]{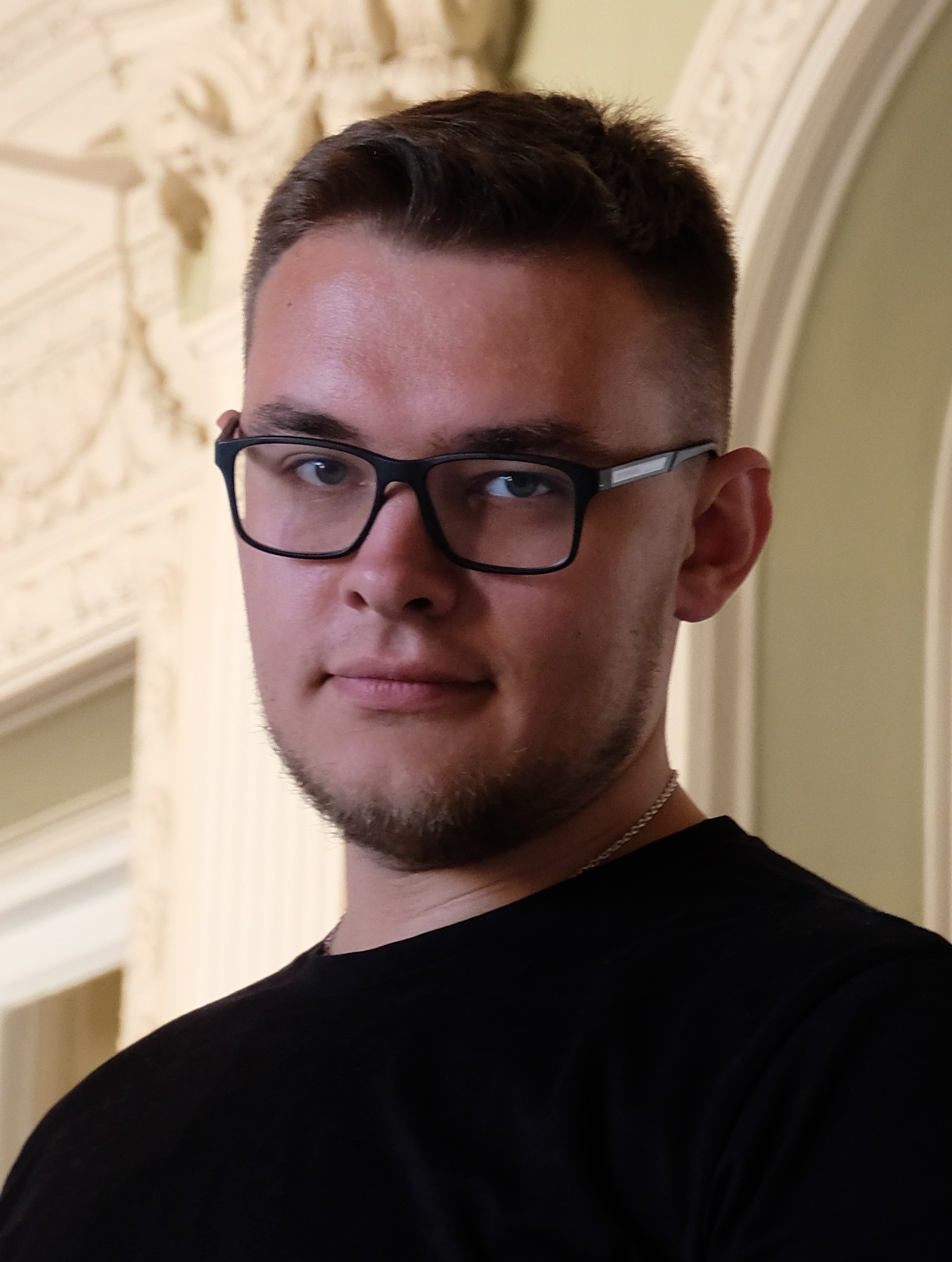}}]{Alexey Rolich} (Member, IEEE) is postdoctoral researcher at the University of Rome La Sapienza. He gained bachelor’s and master’s degrees at National Research University Higher School of Economics in Moscow, Russia in 2013 and 2015. He gained PhD degree at University of Rome La Sapienza, in 2025. Research interests are focused on vehicular networks, medium access control protocols, cellular radio networks, simulations and performance evaluation.
\end{IEEEbiography}

\begin{IEEEbiography}[{\includegraphics[width=1in,height=1.25in,clip,keepaspectratio]{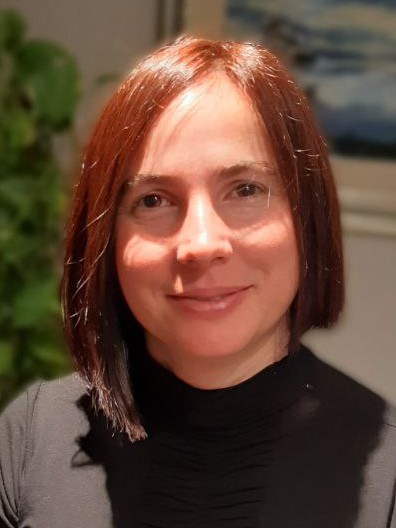}}]{Barbara Bruno} joined the Karlsruhe Institute of Technology
in May 2023 as a Tenure Track Assistant Professor at the Institute
for Anthropomatics and Robotics.
She holds a M.Sc. degree and Ph.D. degree in Robotics
from the University of Genoa, Italy. Upon completion of her
PhD, she co-founded the start-up company Teseo, Italy, focusing on assistive technologies for older adults. In 2019-2023, she was a Postdoctoral Researcher and lab deputy head at the École Polytechnique Fédérale de Lausanne (EPFL), Switzerland.  Her research interests lie
in Socially Assistive Robotics and Human-Robot Interaction.
\end{IEEEbiography}

\begin{IEEEbiography}[{\includegraphics[width=1in,height=1.25in,clip,keepaspectratio]{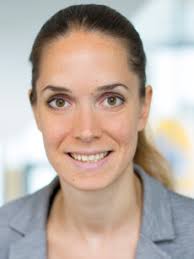}}]{Maike Schwammberger} is an Assistant Professor at Karlsruhe Institute of Technology (KIT), Germany. She received her doctoral degree in computer science from Carl von Ossietzky University Oldenburg. Her research interests include the specification and analysis of various aspects of autonomous traffic maneuvers and making autonomous systems more understandable through self-explainability by generating and verifying explanations from system specifications. 

\end{IEEEbiography}

\begin{IEEEbiography}[{\includegraphics[width=1in,height=1.25in,clip,keepaspectratio]{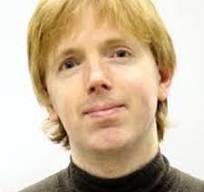}}]{Dieter Fiems} (Senior Member, IEEE) is a Full Professor with the Department of Telecommunications and Information Processing, Ghent University, Belgium. He received his Ph.D. degree in engineering from Ghent University in 2004. His current research interests include various applications of stochastic processes for the performance analysis of communication networks. Particularly, he is interested in applications of queuing theory and branching processes in wireless networks.
\end{IEEEbiography}

\begin{IEEEbiography}[{\includegraphics[width=1in,height=1.25in,clip,keepaspectratio]{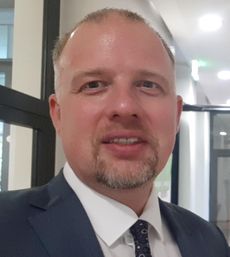}}]{Alexey Vinel} (Senior Member, IEEE) is a Full Professor with Karlsruhe Institute of Technology (KIT), Germany since 2022. Previously, he was a Full Professor with
the University of Passau, Germany. Since 2015,
he has been a Full Professor with Halmstad University, Sweden (currently part-time). He received the Ph.D. degree from Tampere University of Technology, Finland, in 2013. His areas of
interests include vehicular communications and
networking, cooperative autonomous driving, and
future smart mobility solutions.

\end{IEEEbiography}

 




\end{document}